\documentclass[10pt,twocolumn,letterpaper]{article}

\usepackage{iccv}

\usepackage{times}
\usepackage{epsfig}
\usepackage{graphicx}
\usepackage{amsmath}
\usepackage{amssymb}
\usepackage{booktabs}
\usepackage{graphbox} %
\usepackage{capt-of} %
\usepackage{mathtools} %
\usepackage{multirow}
\usepackage{makecell}
\usepackage{placeins} %
\usepackage{adjustbox}
\usepackage{nicefrac}
\usepackage{enumitem}
\usepackage{subcaption}

\newcommand{\myPara}[1]{\noindent\textbf{#1}}

\usepackage[pagebackref=true,breaklinks=true,colorlinks,citecolor=cyan,linkcolor=red,bookmarks=false]{hyperref}

\iccvfinalcopy %

\def\iccvPaperID{} %

\ificcvfinal\fi

\usepackage{pifont}%

\newcommand{\cmark}{\ding{51}}%

\usepackage[capitalize]{cleveref}
\crefname{section}{Sec.}{Secs.}
\Crefname{section}{Section}{Sections}
\Crefname{table}{Table}{Tables}
\crefname{table}{Tab.}{Tabs.}

\graphicspath{ {./imgs/} {./suppimgs/}  }

\newcommand{\secref}[1]{Sec.~\ref{#1}}
\newcommand{\tabref}[1]{Table.~\ref{#1}}
\newcommand{\figref}[1]{Fig.~\ref{#1}}
\renewcommand{\eqref}[1]{Eq.~\ref{#1}}

\newcommand{\RR}{\mathbb{R}}
\newcommand{\codebook}{\mathcal{Z}}

\newcommand{\decoder}{G}
\newcommand{\encoder}{E}
\newcommand{\quantize}{\mathbf{q}}
\newcommand{\quantizedcode}{z_{\mathbf{q}}}
\begin{document}

\title{Rethinking the Objectives of Vector-Quantized Tokenizers for Image Synthesis}

\author{Yuchao Gu$^{1}$,\; Xintao Wang$^{2}$,\; Yixiao Ge$^{2}$,\; Ying Shan$^{2}$,\; Xiaohu Qie$^{3}$,\; Mike Zheng Shou$^1$\thanks{Corresponding Author.}\\ \\
$^1$Show Lab, National University of Singapore\quad $^2$ARC Lab,$^3$Tencent PCG \\
\url{https://github.com/TencentARC/BasicVQ-GEN}
}

\maketitle

\begin{abstract}
Vector-Quantized (VQ-based) generative models usually consist of two basic components, \textit{i.e.}, VQ tokenizers and generative transformers. Prior research focuses on improving the reconstruction fidelity of VQ tokenizers but rarely examines how the improvement in reconstruction affects the generation ability of generative transformers. In this paper, we surprisingly find that improving the reconstruction fidelity of VQ tokenizers does not necessarily improve the generation. Instead, learning to compress semantic features within VQ tokenizers significantly improves generative transformers' ability to capture textures and structures. We thus highlight two competing objectives of VQ tokenizers for image synthesis: \textbf{semantic compression} and \textbf{details preservation}. Different from previous work that prioritizes better details preservation, we propose \textbf{Se}mantic-\textbf{Q}uantized GAN (SeQ-GAN) with two learning phases to balance the two objectives. In the first phase, we propose a semantic-enhanced perceptual loss for better semantic compression. In the second phase, we fix the encoder and codebook, but enhance and finetune the decoder to achieve better details preservation. 
Our proposed SeQ-GAN significantly improves VQ-based generative models for both unconditional and conditional image generation. Specifically, SeQ-GAN achieves a Fréchet Inception Distance (FID) of 6.25 and Inception Score (IS) of 140.9 on 256×256 ImageNet generation, which is a remarkable improvement over VIT-VQGAN (714M), which obtains 11.2 FID and 97.2 IS.
\end{abstract}

\section{Introduction}
\label{sec:intro}

\begin{figure}[!tb]
    \centering
    \includegraphics[width=\linewidth]{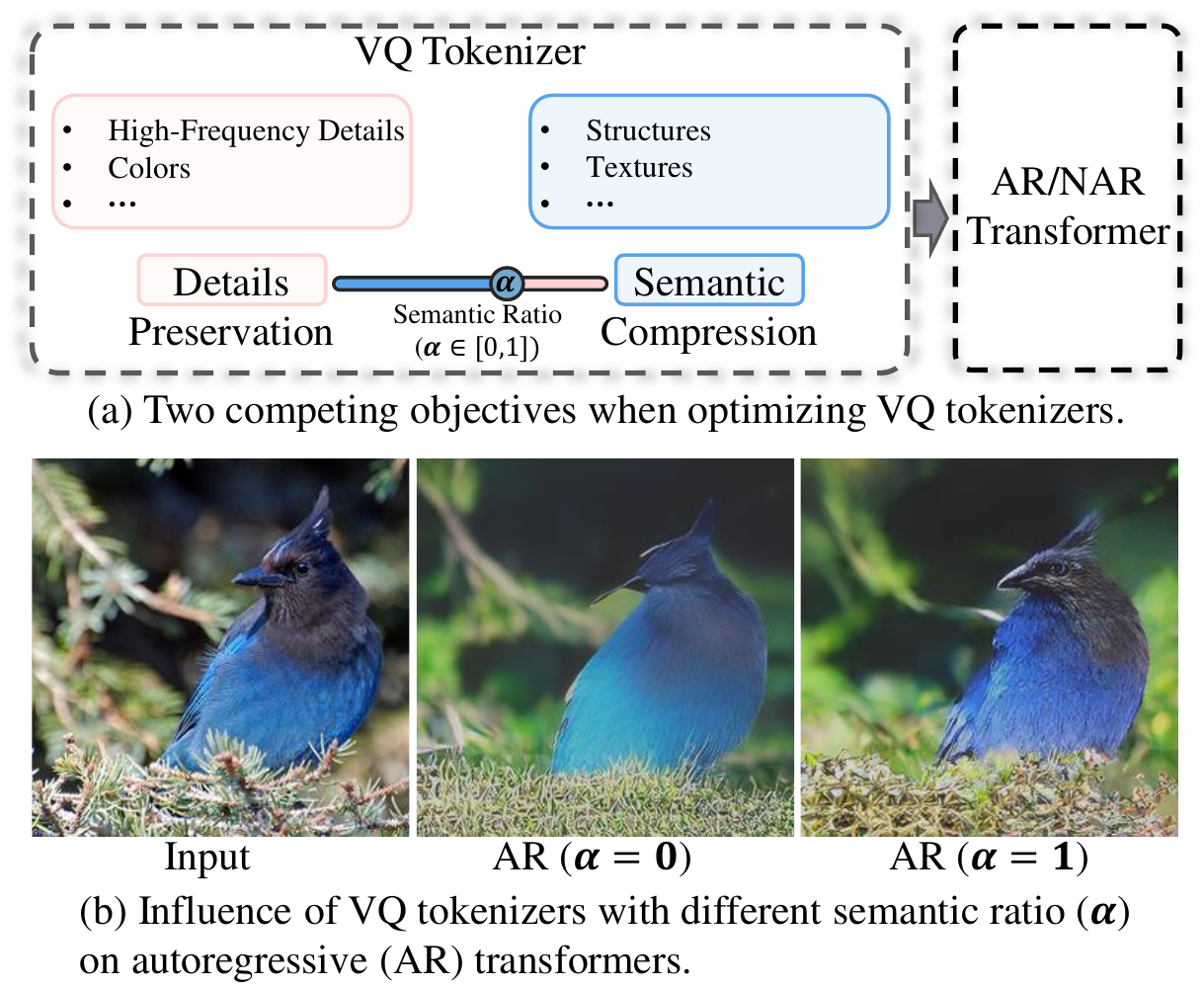}
    \vspace{-.26in}
    \caption{\label{fig:teaser}
    Visualizing impact of VQ tokenizers on generative transformers with $\alpha$ trade-off between details preservation and semantic compression in VQ tokenizer training.}
    \vspace{-.2in}
\end{figure}

In recent years, remarkable progress has been made in image synthesis using likelihood-based generative methods, such as diffusion models~\cite{dhariwal2021diffusion,rombach2022high}, autoregressive (AR)~\cite{esser2021taming, yu2021vector,yu2022scaling,ramesh2021zero}, and non-autoregressive (NAR)~\cite{chang2022maskgit,gu2022vector} transformers. These models offer stable training and better diversity compared to Generative Adversarial Networks (GANs)~\cite{karras2019style,karras2020analyzing}. However, unlike GANs, which can generate high-resolution (\textit{e.g.}, 256$^2$ and 512$^2$) images at one forward pass, likelihood-based methods usually require multiple forward passes by sequential decoding~\cite{esser2021taming,yu2021vector} or iterative refinement~\cite{chang2022maskgit,gu2022vector}.
Consequently, early works~\cite{chen2020generative,parmar2018image,ho2020denoising}, which maximize likelihood on pixel space, are limited in their ability to synthesize high-resolution images due to the high computational cost and slow decoding speed.

\begin{figure*}[!tb]
    \centering
    \includegraphics[width=\linewidth]{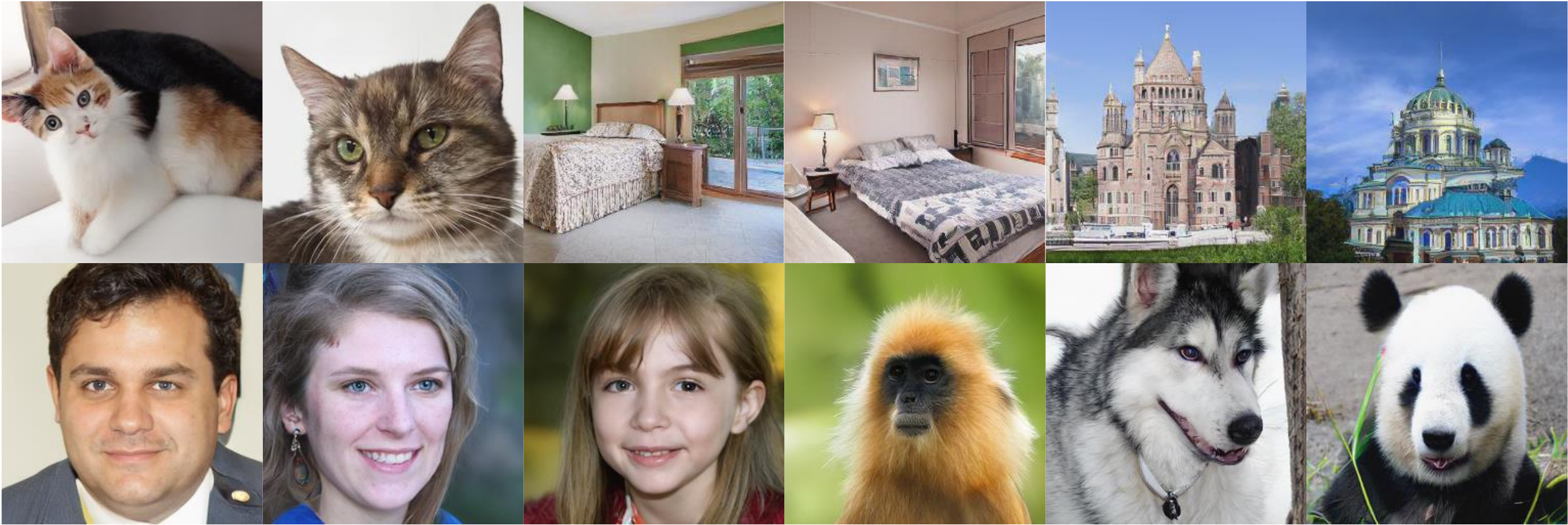}
    \vspace{-.3in}
    \caption{\label{fig:vis_results}
   Generation results of SeQ-GAN+NAR. 1st row: LSUN-\{cat, bedroom, church\}. 2nd row: FFHQ and ImageNet.}
   \vspace{-.22in}
\end{figure*}

Instead of directly modeling the underlying distribution in the pixel space, recent vector-quantized (VQ-based) generative models~\cite{van2017neural} construct a discrete latent space for generative transformers.
There are two basic components in VQ-based generative models, \textit{i.e.}, VQ tokenizers and generative transformers.
VQ tokenizers learn to quantize images into discrete codes, and then decode the codes to recover the input images, which process is termed as \textbf{\textit{reconstruction}}. 
Then, a generative transformer is trained to learn the underlying distribution in the discrete latent space constructed by the VQ tokenizer. Once trained, the generative transformer can be used to sample images from the underlying distribution, and this process is termed as \textbf{\textit{generation}}.
Thanks to the discrete latent space, VQ-based generative models~\cite{esser2021taming, chang2022maskgit,rombach2022high} can easily scale up to synthesize high-resolution images without prohibitive computation cost. 

The VQ tokenizer has received much attention as the core component in VQ-based generative models.
Various techniques, such as  factorized codes and smaller compression ratio in VIT-VQGAN~\cite{yu2021vector}, recursive quantization in Residual Quantization~\cite{lee2022autoregressive}, and multichannel quantization with spatial modulated decoder in MoVQ~\cite{zheng2022movq}, have been used to compress more fine-grained details into VQ tokenizers, leading to steadily improving reconstruction fidelity.
However, none of the previous works have carefully examined a fundamental
question, \textbf{\textit{how the improved reconstruction of VQ tokenizers affects the generation}}. Lacking such analysis is due to two main reasons: 1) the underlying assumption that ``\textit{better reconstruction, better generation}", and
2) the absence of a visualization pipeline to intuitively compare generation results of various VQ tokenizers.

In this paper, we introduce a visualization pipeline for examining how different VQ tokenizers influence generative transformers.
Unlike previous works that compare randomly-sampled generation results, our approach models specific images and facilitates a straightforward comparison of the generative transformer's ability using different VQ tokenizers. 
The key idea is to reduce the flexibility of the sampling process by providing ground-truth context to generative transformers, which can be easily implemented with an autoregressive (AR) transformer with causal attention.

Our proposed visualization pipeline leads us to two important observations. 
1) Improving the reconstruction fidelity of VQ tokenizers does not
necessarily improve the generation. 2) Learning to compress semantic features within VQ tokenizers significantly improves generative transformers’ ability to capture textures and structures.
As shown in \figref{fig:teaser}, increasing the semantic ratio ($\alpha$=1) improves the AR transformer's ability to capture texture and structure, while decreasing it ($\alpha$=0) results in the transformer modeling rough colors instead.
These observations arise due to the competing objectives of reconstruction and generation optimization. Reconstruction aims to retain variation in the dataset by favoring latent spaces with larger variance ($\textit{i.e.}$, weaker separability), whereas generation optimization favors latent spaces with smaller variance ($\textit{i.e.}$, better separability) to optimize a classification objective.

Our observations reveal that there are two competing objectives for VQ tokenizers: \textbf{\textit{semantic compression}} and \textbf{\textit{details preservation}}, but recent VQ tokenizers~\cite{razavi2019generating,zheng2022movq,lee2022autoregressive,yu2021vector} have primarily focused on the latter.
To balance the two objectives for better generation, 
we propose \textbf{Se}mantic-\textbf{Q}uantized GAN (SeQ-GAN), which consists of two learning phases. The first phase utilizes a semantic-enhanced perceptual loss to achieve semantic compression, while the second phase finetunes the decoder to restore fine-grained details while preserving structures and textures. Compared to previous VQ tokenizers, SeQ-GAN compresses semantic features rather than fine-grained details (\textit{e.g.}, high-frequency details, colors) into codebook and finetunes the decoder to restore those details, which does not affect transformer learning but improves local details generation.

Our main contributions are summarized as follows. (1) We rethink the common assumption "\textit{better reconstruction, better generation}" in recent VQ tokenizers, and propose a visualization pipeline to explore the impact of different VQ tokenizers on generative transformers.
(2) We identify two competing objectives in optimizing VQ tokenizers: \textbf{\textit{semantic compression}} and \textbf{\textit{details preservation}}, and introduce SeQ-GAN as a solution that balances these objectives to achieve better generation quality.
(3) Our SeQ-GAN achieves significant improvements over prior VQ tokenizers in both conditional and unconditional image generation, as demonstrated through experiments with both AR and NAR transformers.
(Generation results are shown in \figref{fig:vis_results}).
\section{Related Work}
\label{sec:related}
\myPara{VQ-based Generative Models.}
The VQ-based generative model is first introduced by VQ-VAE~\cite{van2017neural}, which constructs a discrete latent space by VQ tokenizers and learns the underlying latent distribution by prior models~\cite{van2016conditional,chen2018pixelsnail}. VQGAN~\cite{esser2021taming} improves upon this by utilizing perceptual loss~\cite{zhang2018unreasonable,johnson2016perceptual} and adversarial learning~\cite{goodfellow2020generative} in training VQ tokenizers, and using autoregressive transformers~\cite{radford2019language} as the prior model, leading to significant improvements in generation quality. VQ-based generative models have been applied in various generation tasks, such as image generation~\cite{yu2021vector,chang2022maskgit,esser2021taming}, video generation~\cite{ge2022long,yan2021videogpt,hong2022cogvideo}, text-to-image generation~\cite{ramesh2021zero,ding2021cogview,rombach2022high,yu2022scaling}, and face restoration~\cite{gu2022vqfr,wang2022restoreformer,zhou2022towards}.

Building on the success of VQGAN~\cite{esser2021taming}, recent works have focused on improving the two fundamental components of VQ-based generative models: VQ tokenizers and generative transformers. To enhance VQ tokenizers, VIT-VQGAN~\cite{yu2021vector} proposes quantizing image features into factorized and L2-normed codes with a larger codebook and small compression ratio, achieving finer reconstruction results. Residual Quantization~\cite{lee2022autoregressive} recursively quantizes feature maps using a shared codebook to precisely approximate image features. MoVQ~\cite{zheng2022movq} enhances the VQ tokenizer's decoder with modulation~\cite{huang2017arbitrary} and proposes multi-channel quantization with a shared codebook, resulting in state-of-the-art reconstruction results.
Different from previous works, we argue that improving reconstruction fidelity does not necessarily lead to better generation quality.

Another line orthogonal to our work is improving generative transformers. Early works adopt autoregressive (AR) transformers~\cite{esser2021taming,yu2021vector,razavi2019generating}. However, AR transformers suffer from low sampling speed and ignore bidirectional contexts. To overcome these limitations, non-autoregressive (NAR) transformers are introduced based on different theories, like mask image modeling~\cite{bao2021beit,he2022masked} (\textit{i.e.}, MaskGIT~\cite{chang2022maskgit}) and discrete diffusion~\cite{austin2021structured,hoogeboom2021argmax} (\textit{i.e.}, VQ-diffusion~\cite{gu2022vector,tang2022improved}). 
In this paper, we demonstrate that integrating our SeQ-GAN as the VQ tokenizer consistently enhances the generation quality of both AR and NAR transformers.

\myPara{Visual Tokenizers for Generative Pretraining.}
Recent works in large-scale generative visual pretraining also explore the potential of the visual tokenizer.
Instead of directly performing mask image modeling on pixels~\cite{he2022masked,xie2022simmim}, the pioneer BEiT~\cite{bao2021beit} reconstructs masked patches quantized by a
discrete VAE~\cite{ramesh2021zero}.
Follow-up works further strengthen the semantics of the visual tokenizer, such as PeCo~\cite{dong2021peco}, which adopts contrastive perceptual loss~\cite{he2020momentum,chen2020improved} during tokenizer training, and mc-BEiT~\cite{li2022mc}, which softens and re-weights the masked prediction target during visual pretraining.
To further reduce the low-level representation in the visual tokenizer,
iBOT~\cite{zhou2021ibot} abandons reconstructing pixels, but updates the tokenizer online during the pretraining. 
BEiT-v2~\cite{peng2022beit} formulates the training objective of the visual tokenizer by reconstructing semantic features extracted by CLIP~\cite{radford2021learning}.
Unlike prior attempts to remove low-level representation interference in visual pretraining, we highlight the importance of semantic compression and details preservation in training VQ tokenizers for image synthesis.

\section{Methodology}
\label{sec:method}

In this section, 
we first review how VQ tokenizers affect generation in VQ-based generative models in \secref{sec:preliminary}. Then, we present a visualization pipeline in \secref{sec:visualization} to examine the impact of different VQ tokenizers on generative transformers. Based on this pipeline, we make two critical observations in \secref{sec:rethink}, highlighting the competing objectives in designing VQ tokenizers. Finally, we propose SeQ-GAN in \secref{sec:SeQ-GAN} as a solution that balances these objectives to improve generation quality.

\subsection{Preliminaries}
\label{sec:preliminary} 
In this section, we cover the fundamental process of VQ-based generative models and highlight the potential impact of VQ tokenizers on generation results.

\begin{figure}[!tb]
    \centering
    \includegraphics[width=0.88\linewidth]{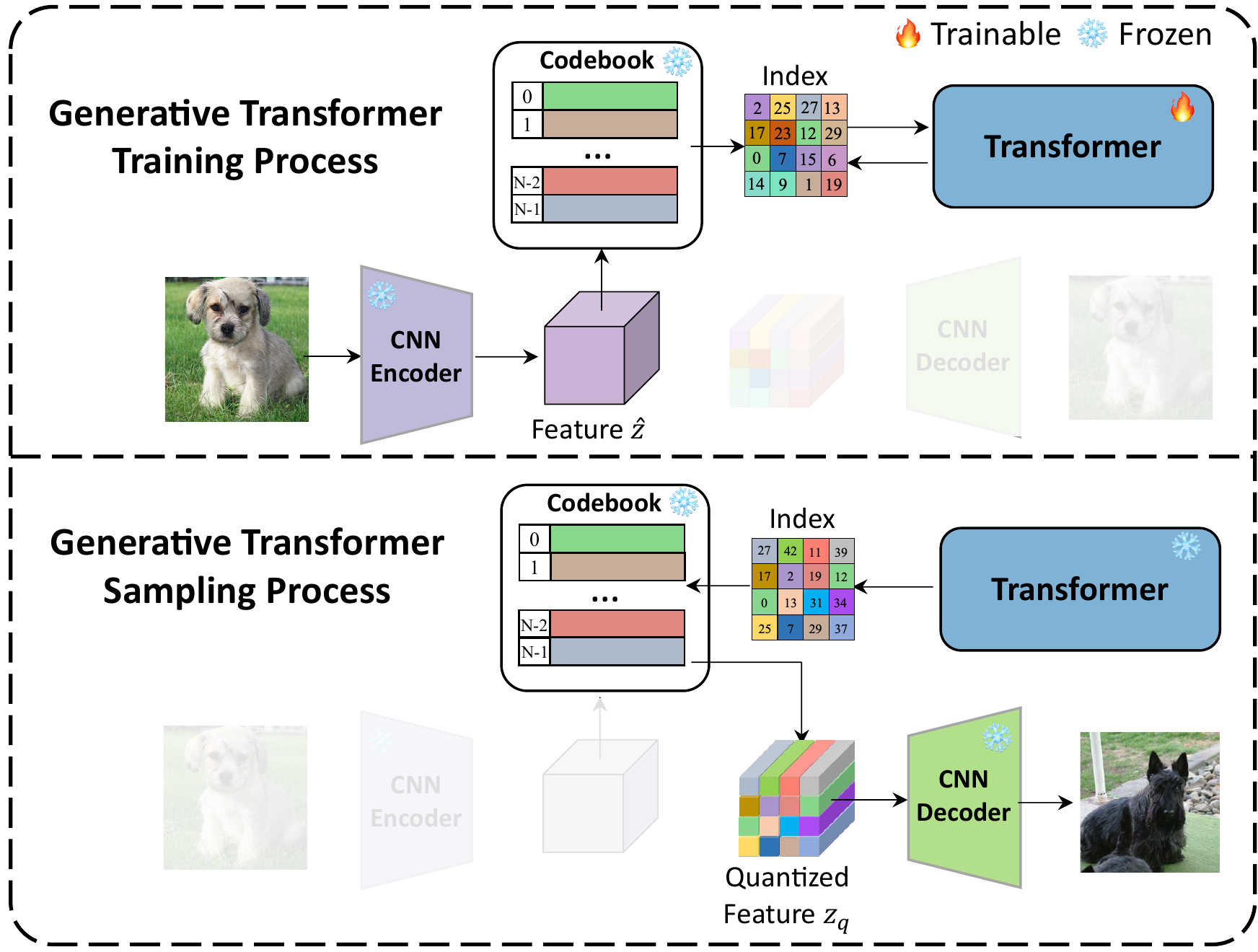}
    \vspace{-.15in}
    \caption{\label{fig:vq_process}
    The influence of VQ tokenizers on the training and sampling process of generative transformers.}
    \vspace{-.2in}
\end{figure}

\myPara{Reconstruction: training VQ tokenizers.}
The role of VQ tokenizers is to compress the image into discrete indices. Specifically, a VQ tokenizer is comprised of an encoder $\encoder$, a decoder $\decoder$ and a codebook $\codebook=\{z_k\}^K_{k=1}$ with $K$ discrete codes.
Given an input image $x\in \RR^{H\times W\times 3}$, a latent feature $\hat{z}\in \RR^{\frac{H}{f}\times \frac{W}{f}\times n_z}$ is first extracted, where $n_z$ and $f$ represent the dimension of the latent features and the spatial compression ratio, respectively. 
Then, the feature vector at each spatial position $(i,j)$ is quantized to the nearest code in the codebook by 
{\setlength{\abovedisplayskip}{1pt}
\setlength{\belowdisplayskip}{1pt}
\begin{equation}
  \quantizedcode = \quantize(\hat{z}) \coloneqq
  \left(\arg\min_{z_k \in \codebook} \Vert \hat{z}_{ij} - z_k \Vert\right)
  \in \RR^{\frac{H}{f}\times \frac{W}{f} \times n_z}.
  \label{eq:vq}
\end{equation}}The decoder $G$ is responsible for decoding the quantized features back to the image space, \textit{i.e.}, $\hat{x} = \decoder(\quantizedcode)$.

The training objective of the VQ tokenizer is to minimize the reconstruction error with respect to the input image.
Following VQGAN~\cite{esser2021taming} to use adversarial loss ($\mathcal{L}_{adv}$)~\cite{goodfellow2020generative} and perceptual loss ($\mathcal{L}_{per}$)~\cite{zhang2018unreasonable,johnson2016perceptual}, the reconstruction objective can be formulated as
{
\setlength{\abovedisplayskip}{1pt}
\setlength{\belowdisplayskip}{1pt}
\begin{equation}
\begin{split}
  \mathcal{L}(&\encoder, \decoder, \codebook) = \mathcal{L}_{vq} + \mathcal{L}_{per} + \mathcal{L}_{adv}, where\\
  \mathcal{L}_{vq}=\Vert x - &\hat{x} \Vert_1 
  + \Vert \text{sg}[\encoder(x)] - \quantizedcode \Vert_2^2 + \beta\Vert \text{sg}[\quantizedcode] - \encoder(x) \Vert_2^2.
\end{split}
 \label{eq:vqobjective}
\end{equation}
}In \eqref{eq:vqobjective}, $\text{sg}[\cdot]$ means stop-gradient and $\beta\Vert \text{sg}[\quantizedcode] - \encoder(x) \Vert_2^2$ is known as the commitment loss~\cite{van2017neural}, where the commitment weight $\beta$ is set to 0.25 following ~\cite{van2017neural,esser2021taming,yu2021vector}.

\myPara{Training generative transformers.}
As shown in \figref{fig:vq_process}, the encoder and codebook of a trained VQ tokenizer define a discrete latent space that quantizes an image into a sequence of discrete indices for generative transformer training. This sequence serves as input and label in training the generative transformer with token classification loss. In this paper, we use the autoregressive (AR) transformer in VQGAN~\cite{esser2021taming} and the non-autoregressive (NAR) transformer in MaskGIT~\cite{chang2022maskgit}. Therefore, the quality of the discrete latent space defined by the encoder and codebook of VQ tokenizers will influence the generative transformer training.

\myPara{Generation: sampling from generative transformers.}
After training a generative transformer, we can sample discrete index sequences from it through either autoregressive decoding~\cite{esser2021taming} or iterative refinement~\cite{chang2022maskgit}. To map the discrete indices back to visual details, we retrieve the corresponding feature from the codebook and decode it into image space using the VQ tokenizer's decoder. Therefore, the decoder will affect the generation quality by influencing the index-to-visual-details mapping.

\begin{figure}[!tb]
    \centering
    \includegraphics[width=\linewidth]{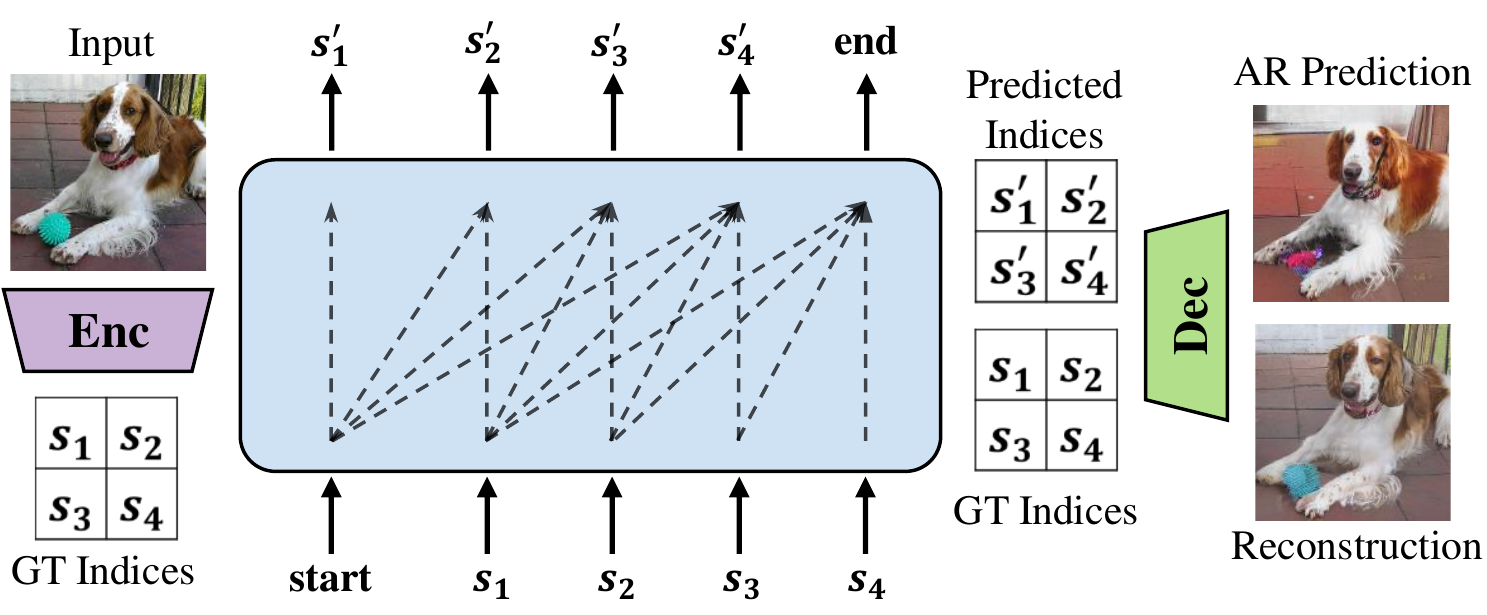}
    \vspace{-.3in}
    \caption{\label{fig:analysis_tool}Visualization pipeline to examine the influence of VQ tokenizers on generative transformers.}
    \vspace{-.2in}
\end{figure}

\subsection{Pipeline for Visualizing VQ Generative Models}
\label{sec:visualization}
Recent VQ-based generative models examine their designs by looking into the random sampled generation results, where different sampling techniques are adopted (\textit{e.g.}, top-$p$ top-$k$ sampling~\cite{holtzman2019curious}, classifier-free guidance~\cite{ho2022classifier}, or rejection sampling~\cite{ramesh2021zero}). 
However, instead of examining random samples, we are more curious about how generative transformers model specific images, enabling us to check the influence of different VQ tokenizers on generative transformers side by side. To achieve that goal, we propose to reduce the flexibility of the sampling process by providing ground-truth (GT) contexts for predicting each index, which can be easily implemented by AR transformers.

\begin{table}[!tb]
\setlength\tabcolsep{3pt}
\centering
\small
\begin{tabular}{l|c|c|cccc}
\toprule
\multirow{2}{*}{Model}        & \multirow{2}{*}{Params} & \multirow{2}{*}{rFID$\downarrow$} & \multicolumn{4}{c}{Generation FID$\downarrow$}   \\
        &  &  & AR & AR-L & AR-L-2$\times$ & NAR   \\\midrule
baselineVQ   & 54.5M    & 3.45 & \textbf{16.97} & \textbf{13.86} & \textbf{11.49} & \textbf{13.26} \\
+Conv$\times 2$      & 70.0M    & 3.22 & 17.19 & 14.50 & 12.03 & 13.51 \\
+Attention$\times 2$ & 61.4M    & \textbf{2.90} & 17.42 & 14.91 & 12.04 & 14.02 \\ \bottomrule
\end{tabular}
\vspace{-.13in}
\caption{\label{tab:motivation}
Comparison of the baseline and  decoder-enhanced VQ tokenizers on the reconstruction FID (rFID) and generation FID, evaluated on different transformer configurations.}
\vspace{-.2in}
\end{table}

The pipeline is shown in \figref{fig:analysis_tool}. First, we train a VQ tokenizer along with its corresponding AR transformer.
To analyze a specific image, we obtain the GT index sequence $s=[s_i]_{i=1}^{N}$ from the VQ tokenizer and feed it to the trained AR transformer, similar to the teacher forcing strategy~\cite{bengio2015scheduled,williams1989learning} used in training AR transformers.
Because the AR transformer adopts casual attention~\cite{radford2019language}, it does not directly access the GT indices, but can accesses all GT context indices for predicting each index. Given the same context (\textit{i.e.}, preceding GT indices), the next index prediction task is well-controlled and thus we can get the top-1 predicted index sequence $s'$ within one forward pass.
Finally, we decode the GT sequence $s$ and the AR predicted sequence $s'$ back to the image space by the decoder of the VQ tokenizer.
Following this approach, we are able to visualize both the reconstruction of VQ tokenizers and the upper limit prediction of AR transformers for specific images. 

\subsection{Rethinking the Objectives of VQ Tokenizers}
\label{sec:rethink}

\subsubsection{Reconstruction \textit{vs.} Generation}
\label{sec:observation1}
\noindent\textbf{Motivation.} 
Recent advancements in VQ tokenizers have led to improved reconstruction results, with MoVQ~\cite{zheng2022movq} in particular enhancing their decoder with modulation to add variation to quantized code and achieve the highest reconstruction fidelity. However, few studies investigate whether improvements of reconstruction fidelity of VQ tokenizers benefit generation quality. To address this gap, we conduct the following experiment to answer this question.

\noindent\textbf{Experimental Settings.} 
In \secref{sec:preliminary}, we identify two key factors in VQ tokenizers that affect generation: 1) the quality of the discrete latent space defined by the encoder/codebook, and 2) the index-to-visual-details mapping defined by the decoder.
Inspired by MoVQ~\cite{zheng2022movq}, we keep the configuration of encoder/codebook the same, and enhance the decoder to strengthen the index-to-visual-detail mapping. Our baseline is a convolution-only VQGAN~\cite{esser2021taming}, and we add two extra convolution blocks or two interleaved regional and dilated attention blocks~\cite{zhao2021improved} at each resolution level to enhance the decoder. Based on each tokenizer, we train the generative transformer with different configurations, including different parameter sizes (AR and AR-Large), different types (AR and NAR), and different training iterations (AR-Large and AR-Large-2$\times$). Additional experimental settings can be found \textit{in the \secref{sec:setting}}.

\noindent\textbf{Results.}
The results presented in \tabref{tab:motivation} show that enhancing the decoder improves reconstruction fidelity, but it does not necessarily lead to better generation quality. Surprisingly, the baseline tokenizer achieves the best generation quality.
Assuming that the quality of the discrete latent space (defined by encoder/codebook) remains unchanged, enhancing the decoder should improve generation quality by improving the index-visual-details mapping. However, in reality, enhancing the decoder leads to a degradation in generation quality.
This suggests that \textbf{\textit{jointly learning}} the encoder/codebook with an enhanced decoder actually degrades the quality of the discrete latent space.

\begin{figure}[!tb]
    \centering
    \includegraphics[width=0.92\linewidth]{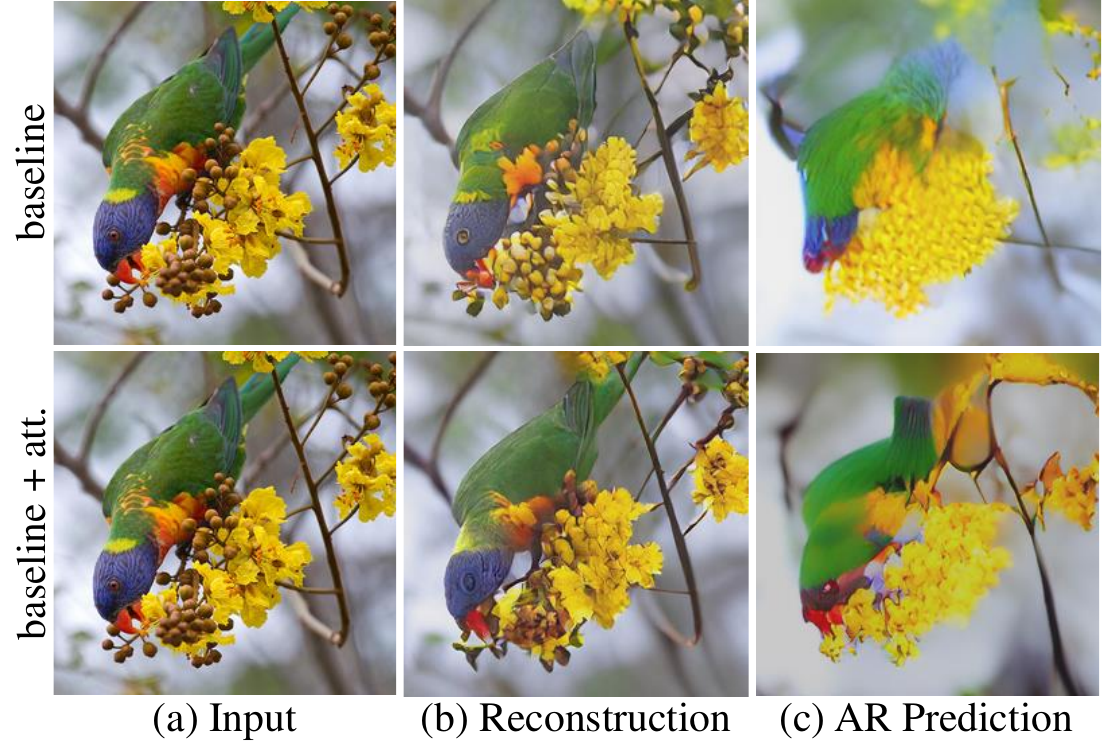}
    \vspace{-.15in}
    \caption{\label{fig:motivation_vis}Visualizing the reconstruction and AR prediction of the baseline tokenizer and its attention-enhanced variant.}
    \vspace{-.22in}
\end{figure}

Using the proposed visualization pipeline in \secref{sec:visualization}, we visualize the reconstruction and AR prediction results of the baseline tokenizer and its attention-enhanced variant in \figref{fig:motivation_vis}.
Although the attention-enhanced variant leads to a more consistent reconstruction, the AR transformer faces challenges in capturing the details and can only predict a rough color for the main object, even given ground-truth contexts. This highlights the generative transformers' difficulties in modeling the discrete latent space.

The discrepancy between reconstruction and generation is due to the conflicting optimization objectives. In reconstruction training, VQ tokenizers prefer a latent space with larger variance (\textit{i.e.}, weaker separability) to retain the variation of datasets, while generative transformer training prefers smaller variance (\textit{i.e.}, better separability), because it optimizes the classification objective (\textit{i.e.}, cross-entropy). Therefore, in \tabref{tab:motivation} and \figref{fig:motivation_vis}, a powerful decoder promotes encoding more variation in the codebook, which hinders the separability of the discrete latent space and thus results in suboptimal generation performance.
Through the result, we arrive at the following observation.

\noindent\textbf{Observation 1.}
\textit{Improving the reconstruction fidelity of VQ tokenizers does not necessarily improve the generation.}

\begin{figure}[!tb]
\centering
\subfloat[Visualization of the AR predicted results when trained using VQ tokenizers with different semantic ratios ($\alpha$).]{
\vspace{-.15in}
\includegraphics[width=0.9\linewidth]{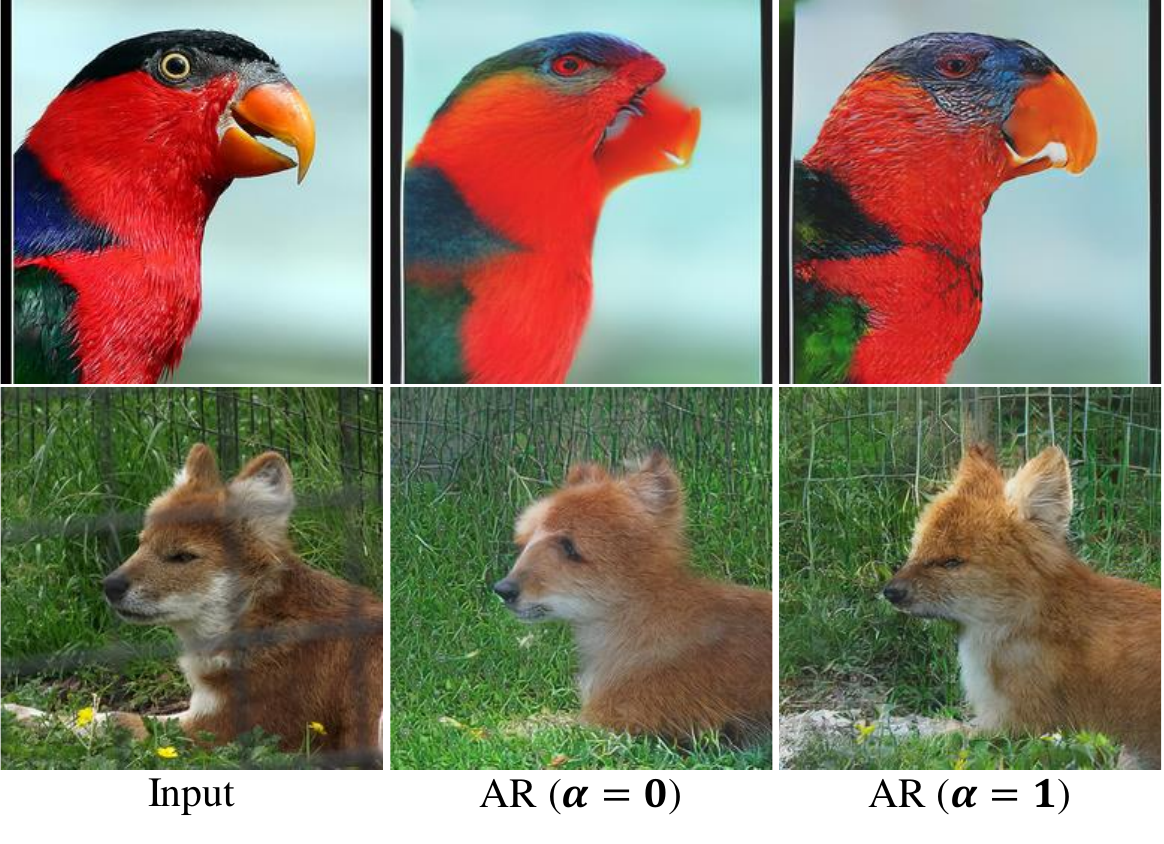}}\\ 
\subfloat[Reconstruction FID and generation FID with different semantic ratio$(\alpha)$ in optimizing VQ tokenizers.]{
\vspace{-.12in}\includegraphics[width=\linewidth]{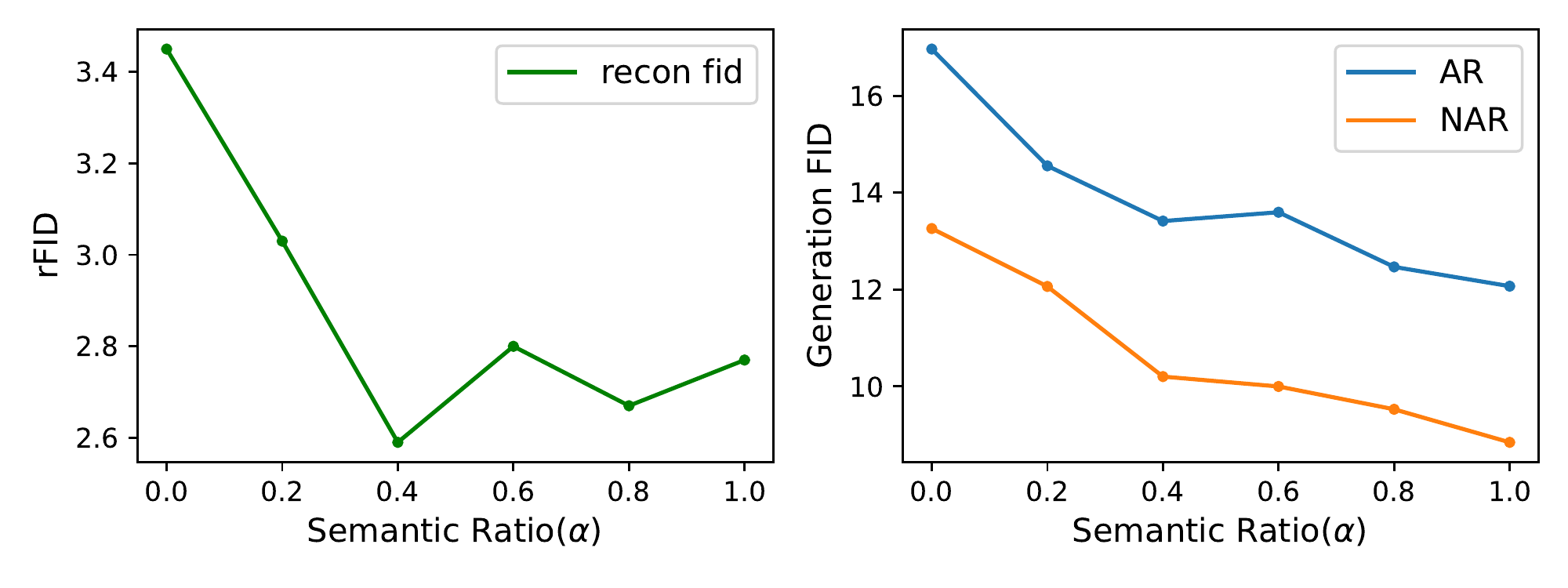}}
\vspace{-.1in}
\caption{\label{fig:semantic_analysis}Influence of the semantic ratio $\alpha$ in VQ tokenizers on generation quality.}
\vspace{-.24in}
\end{figure}

\subsubsection{Details Preservation \textit{vs.} Semantic Compression}

\noindent\textbf{Motivation.} Observation 1 suggests that compressing more fine-grained details within the tokenizer in reconstruction does not always improve generation. Therefore, we shift our focus towards exploring the role of semantics in VQ tokenizers for better generation quality.

\noindent\textbf{Semantic-Enhanced Perceptual Loss.} 
Unlike generative pretraining~\cite{zhou2021ibot,peng2022beit} that uses fully semantic tokenizers, image synthesis requires consideration of low-level details. To balance the trade-off between low-level details and semantics in VQ tokenizers, we introduce a semantic-enhanced perceptual loss that controls the details/semantic ratio.

Specifically, given an input and a reference image, we extract their activation features $\hat{y}^l$ and $y^l$ from a pre-trained VGG~\cite{simonyan2014very} network. 
For each layer $l$, the feature is of shape $H_{l}\times W_{l}\times C_{l}$. Then, the perceptual loss can be calculated as
$
 \mathcal{L}_{per}=\sum_{l}\frac{1}{H_{l}W_{l}C_{l}}||\hat{y}^l-y^l||^2_2.
$
To preserve details, perceptual loss~\cite{zhang2018unreasonable} used in previous VQ tokenizers adopts the features from both the shallow and high layers, which we denote as $L_{per}^{low}$ in this paper. 
To better compress semantic information during reconstruction, we propose a semantic-enhanced perceptual loss $L_{per}^{sem}$, which removes the features from shallow layers and further includes the logit feature (\textit{i.e.}, feature before softmax classifier). The layers $l$ to extract features can be summarized as
\begin{itemize}[itemsep=2pt,topsep=0pt,parsep=0pt]
    \item[-] $\mathcal{L}_{per}^{low}$: $l\in$\{relu-\{1\_2, 2\_2, 3\_3, 4\_3, 5\_3\}\},
    \item[-] $\mathcal{L}_{per}^{sem}$: $l\in$\{relu5\_3, logit\}.
\end{itemize}
We re-weight the two perceptual losses to control the proportion between the details and semantic information by
{\setlength{\abovedisplayskip}{1pt}
\setlength{\belowdisplayskip}{1pt}
\begin{equation}
    \mathcal{L}_{per}^{\alpha}=\alpha\mathcal{L}_{per}^{sem}+(1-\alpha)\mathcal{L}_{per}^{low},
    \label{eq:reweight}
\end{equation}}where $\alpha\in[0,1]$ is the semantic ratio.

\noindent\textbf{Results.} Using the proposed semantic-enhanced perceptual loss, we examine the impact of different semantic ratios ($\alpha$) during VQ tokenizer training on generation quality. In particular, we find that increasing $\alpha$ initially improves the reconstruction FID (rFID), with the best rFID achieved at $\alpha$=0.4 before decreasing. However, increasing $\alpha$ consistently improves the generation FID. 
Our visualizations in \figref{fig:semantic_analysis}(a) and \figref{fig:teaser}(b) demonstrate that the semantic-enhanced VQ tokenizer ($\alpha$=1) enables the AR transformer to capture more overall structures and textures than the baseline tokenizer ($\alpha$=0).
We provide additional visualizations \textit{in the \secref{sec:vis_observation}}. These results help us arrive at the following observation.

\begin{figure}[!tb]
    \centering
    \includegraphics[width=\linewidth]{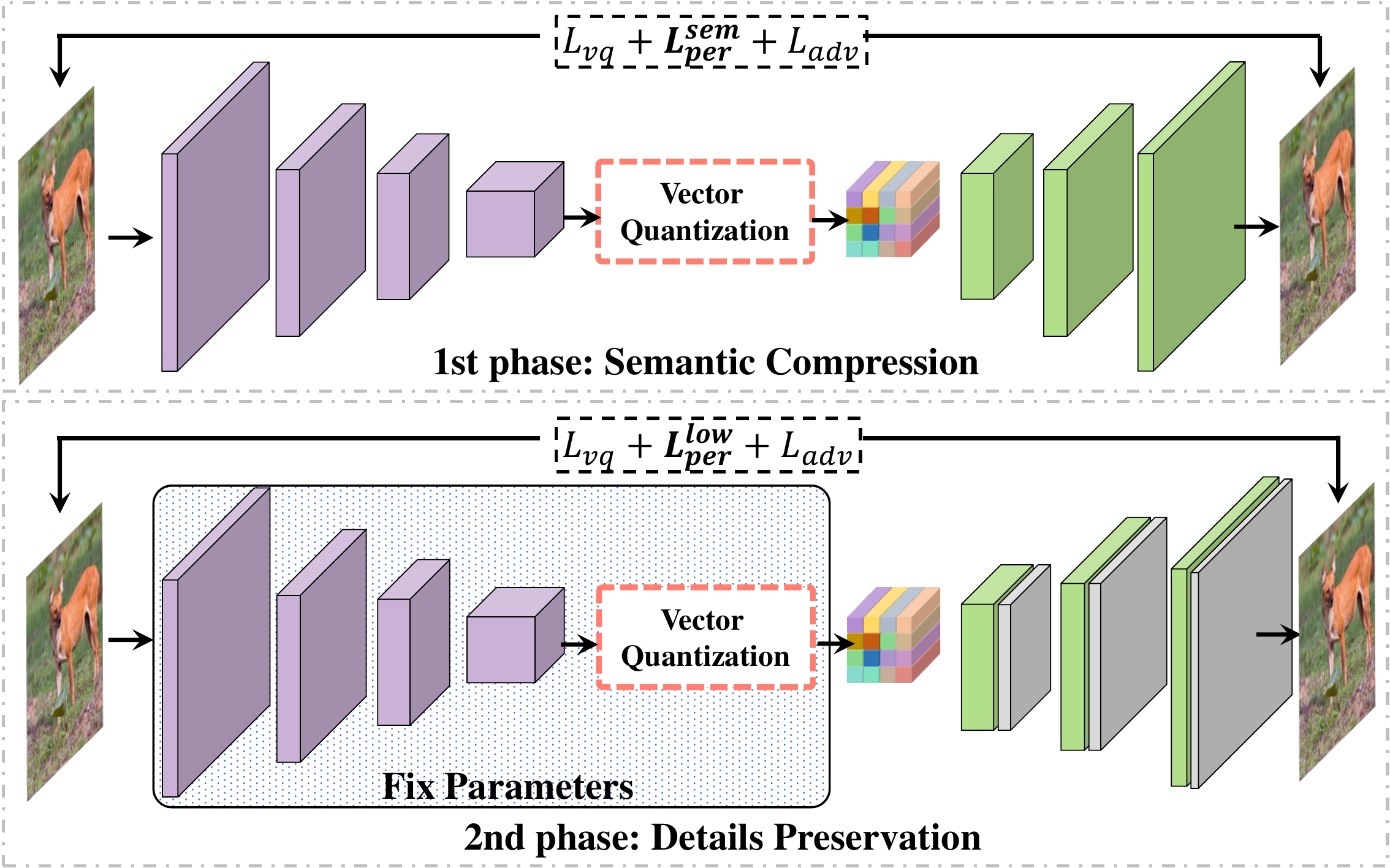}
    \vspace{-.3in}
    \caption{\label{fig:pipeline}
   Pipeline of the two-phase learning in SeQ-GAN. }
   \vspace{-.2in}
\end{figure}

\noindent\textbf{Observation 2.} \textit{Semantic compression within VQ tokenizers benefits the generative transformer.}

\subsubsection{Discussion}
Tokenizers~\cite{kudo2018sentencepiece,sennrich2015neural} in Natural Language Processing (NLP) are naturally discrete and semantically meaningful, and in large-scale generative visual pretraining~\cite{zhou2021ibot,peng2022beit}, fully semantic visual tokenizers that abandon low-level information are preferred. However, VQ tokenizers in VQ-based generative models should consider low-level details. Previous works~\cite{razavi2019generating,zheng2022movq,lee2022autoregressive,yu2021vector} prioritize preserving details to achieve better reconstruction fidelity, but we find solely compressing fine-grained details within VQ tokenizers will degrade the discrete latent space and hinder transformer training. We argue that both semantic compression and detail preservation should be considered when designing VQ tokenizers for image synthesis.

\subsection{Our Solution: SeQ-GAN}
\label{sec:SeQ-GAN}
To achieve better generation quality, we propose the \textbf{Se}mantic-\textbf{Q}uantized GAN (SeQ-GAN) as the VQ tokenizer in VQ-based generative models, balancing the objectives of semantic compression and details preservation.

\figref{fig:pipeline} illustrates the two-phase approach of SeQ-GAN for tokenizer learning. In the first phase, we prioritize semantic compression by applying the proposed semantic-enhanced perceptual loss $\mathcal{L}_{per}^{\alpha=1}$ in \eqref{eq:reweight}. However, semantic compression with VQ tokenizers may cause some loss of color fidelity and high-frequency details. To address this, we enhance the decoder in the second phase using interleaved block regional and dilated attention~\cite{zhao2021improved}. We fix the encoder and codebook of the tokenizer and finetune the enhanced decoder with $\mathcal{L}_{per}^{\alpha=0}$ to achieve better detail preservation. Note that in the second phase of tokenizer learning, we fix the discrete latent space by fixing the encoder and codebook. Therefore, our decoder-only finetuning enhances the generation quality of local details without affecting the transformer learning of structures and textures.

\begin{table}[!tb]
\setlength\tabcolsep{1pt}
\centering
\begin{tabular}{l|c|ccc|c}
\toprule
\multirow{2}{*}{Model} & \multirow{2}{*}{Datset} & Latent & \multicolumn{2}{c|}{Codebook} & \multirow{2}{*}{rFID} \\
& & Size & $K$ & Usage & \\ \midrule
VQGAN~\cite{esser2021taming} & \multirow{5}{*}{FFHQ} & 16$\times$16 & 1024 & 42\% & 4.42 \\
VIT-VQGAN~\cite{yu2021vector} & & 32$\times$32 & 8192 & - & 3.13\\
RQ-VAE~\cite{lee2022autoregressive} & & 16$\times$16$\times$4 & 2048 & - & 3.88 \\ 
MoVQ~\cite{zheng2022movq} & & 16$\times$16$\times$4 & 1024 & - & \textbf{2.26} \\
SeQ-GAN (Ours) & & 16$\times$16 & 1024 & \textbf{100\%} & 3.12 \\ \midrule
VQGAN~\cite{esser2021taming} & \multirow{6}{*}{ImageNet} & 16$\times$16 & 1024 & 44\% & 7.94 \\
VQGAN~\cite{esser2021taming} & & 16$\times$16 & 16384 & 5.9\% & 4.98 \\
VIT-VQGAN~\cite{yu2021vector} & & 32$\times$32 & 8192 & 96\% & 1.28 \\ 
RQ-VAE~\cite{lee2022autoregressive} & & 8$\times$8$\times$16 & 16384 & - & 1.83 \\ 
MoVQ~\cite{zheng2022movq} & & 16$\times$16$\times$4 & 1024 & - & \textbf{1.12} \\
SeQ-GAN (Ours) & & 16$\times$16 & 1024 & \textbf{100\%} & 1.99 \\ 
\bottomrule
\end{tabular}
\vspace{-.15in}
\caption{\label{tab:codebook}
Reconstruction results on ImageNet and FFHQ validation set, with $K$ representing codebook size.}
\vspace{-.2in}
\end{table}

\begin{table*}[!tb]
\setlength\tabcolsep{5pt}
\centering
\begin{tabular}{l|c|c|cccc}
\toprule
Model & Params & steps & FFHQ & Church & Cat  & Bedroom \\ \midrule
BigGAN~\cite{brock2018large} & 164M & 1 & 12.4 & - & - & - \\
StyleGAN2~\cite{karras2020analyzing} & 30M & 1 & 3.8 & 3.86 & 7.25 & 2.35 \\
ADM~\cite{dhariwal2021diffusion} & 552M & 1000 & - & - & 5.57 & 1.90 \\
DDPM~\cite{ho2020denoising} & 114M$^{\dag}$/256M$^{\ddag}$ & 1000 & - & 7.89$^{\dag}$ & 19.75$^{\dag}$ & 4.90$^{\ddag}$ \\ 
DCT~\cite{nash2021generating} & 473M$^{\dag}$/448M$^{\ddag}$ & \textgreater{}1024 & 13.06$^{\dag}$ & 7.56$^{\ddag}$ & - & 6.40$^{\ddag}$\\ \midrule
VQGAN~\cite{esser2021taming} & 72.1M + 801M & 256 & 11.4 & 7.81 & 17.31 & 6.35 \\
ImageBART~\cite{esser2021imagebart} & - & - & 9.57 & 7.32 & 15.09 & 5.51 \\
VIT-VQGAN~\cite{yu2021vector} & 64M + 1697M & 1024 & 5.3 & - & - & - \\
RQ-VAE~\cite{lee2022autoregressive} & 100M + 370M$^{\dag}$/650M$^{\ddag}$ & 256 & 10.38$^{\dag}$ & 7.45$^{\dag}$ & 8.64$^{\ddag}$ & 3.04$^{\ddag}$ \\
MoVQ + AR~\cite{zheng2022movq} & 82.7M + 307M & 1024 & 8.52 & - & - & - \\
MoVQ + NAR~\cite{zheng2022movq} & 82.7M + 307M & 12 & 8.78 & - & - & - \\ \midrule
SeQ-GAN + AR (Ours) & 57.9M + 171M & 256 & - & 2.45 & \textbf{3.61} & \textbf{1.44} \\
SeQ-GAN + NAR (Ours) & 57.9M + 171M & 12 & \textbf{3.62} & \textbf{2.25} & 4.60 & 2.05 \\ \bottomrule
\end{tabular}
\vspace{-.14in}
\caption{\label{tab:uncondition}
Quantitative comparison of unconditional image generation on FFHQ~\cite{karras2019style} and LSUN~\cite{yu2015lsun}-\{Church, Cat, Bedroom\}. The AR transformer result on FFHQ is omitted due to severe overfitting, consistent with findings in RQ-VAE~\cite{lee2022autoregressive}.}
\vspace{-.18in}
\end{table*}

During the training of our SeQ-GAN, we observe the issue of low codebook usage, which has also been reported in prior VQGAN~\cite{esser2021taming}. To address this issue, we incorporate entropy regularization techniques that are commonly used in self-supervised representation learning~\cite{li2021contrastive,caron2020unsupervised} to mitigate the problem of empty clusters.
Specifically, given the feature before quantization $\hat{z}\in\RR^{N\times n_z}$, we aims to map $\hat{z}$ to the codebook feature $\codebook=\{z_k\}^K_{k=1}$. Denote the matrix $\mathcal{D}\in\RR^{N\times K}$ as the L2 distance between each feature $\hat{z}_i$ and each code entry $z_k$, we normalize it by softmax $\mathcal{D}_{i,k}=\frac{\exp(-\mathcal{D}_{i,k})}{\sum_{k=1}^{K}\exp(-\mathcal{D}_{i,k})}$. Then we average $\mathcal{D}$ along the spatial size by $\Bar{\mathcal{D}_k}=\frac{1}{N}\sum_{i=1}^{N}\mathcal{D}_{i,k}$, where $\Bar{\mathcal{D}}\in \RR^{K}$ can be interpreted as a soft codebook usage. To increase the codebook usage, we encourage a smoother $\Bar{\mathcal{D}}$, which achieved by penalizing the entropy $H(\Bar{\mathcal{D}})=-\sum_{k}\Bar{\mathcal{D}}_{k}\log \Bar{\mathcal{D}}_{k}$. 
And we update the optimization objective in \eqref{eq:vqobjective} to $\mathcal{L}_{vq'}=\mathcal{L}_{vq} + \gamma H(\Bar{\mathcal{D}})$, where we fix $\gamma=0.01$ in our experiments.
\section{Experiments}

\subsection{Image Quantization}
We train the SeQ-GAN on ImageNet~\cite{deng2009imagenet},  FFHQ~\cite{karras2019style} and LSUN~\cite{yu2015lsun}, separately. In the first phase, we train the SeQ-GAN on ImageNet and FFHQ using the Adam~\cite{kingma2014adam} optimizer with a learning rate of 1e-4 for 500,000 iterations. 
For LSUN-\{cat, bedroom, church\}, we follow RQ-VAE~\cite{lee2022autoregressive} to use the pretrained SeQ-GAN on ImageNet and finetune for one epoch on each dataset. 
In the second phase, we finetune the enhanced decoder of SeQ-GAN on three datasets for 200,000 iterations with a learning rate of 5e-5. Detailed settings are provided \textit{in the \secref{sec:setting}}.

The results are summarized in \tabref{tab:codebook}. 
Since VIT-VQGAN~\cite{yu2021vector}, RQ-VAE~\cite{lee2022autoregressive} and MoVQ~\cite{zheng2022movq} prioritize the reconstruction fidelity by compressing more fine-grained details within the tokenizer,
they usually require a larger latent size. Our SeQ-GAN does not pursue the reconstruction fidelity, but optimizes for better generation quality. Therefore, SeQ-GAN does not achieve the best reconstruction fidelity. However, compared to VQGAN~\cite{esser2021taming}, with the same latent size and codebook size, SeQ-GAN still has a large improvement in rFID and codebook usage.

\begin{table}[!tb]
\setlength\tabcolsep{3pt}
\centering
\begin{tabular}{l|c|c|cc}
\toprule
Model & Params & Steps & FID & IS \\ \midrule
BigGAN-Deep~\cite{brock2018large} & 160M & 1 & 6.95  & 198.2 \\
DCT~\cite{nash2021generating} & 738M & \textgreater{}1024 & 36.51 & - \\
Improved DDPM~\cite{nichol2021improved} & 280M & 250 & 12.26 & - \\
ADM~\cite{dhariwal2021diffusion} & 554M & 250 & 10.94 & 101.0 \\\midrule\midrule
VQ-VAE-2~\cite{razavi2019generating} & 13.5B & 5120 & 31.11 & $\sim$45\\
VQGAN~\cite{esser2021taming} & 1.4B & 256 & 15.78 & 78.3\\
VIT-VQGAN~\cite{yu2021vector} & 714M & 256 & 11.20 & 97.2\\
VIT-VQGAN~\cite{yu2021vector} & 1.7B & 1024 & 4.17 & 175.1 \\
RQ-VAE~\cite{lee2022autoregressive} & 1.4B & 1024 & 8.71 & 119.0 \\
MoVQ + AR~\cite{zheng2022movq} & 389M & 1024 & 7.13  & 138.3 \\ \midrule
SeQ-GAN + AR & 229M & 256 & 7.55 & 121.3 \\
SeQ-GAN + AR-L & 364M & 256 & \textbf{6.25} & \textbf{140.9} \\ \midrule\midrule
VQ-Diffusion~\cite{gu2022vector} & 370M & 100 & 11.89 & - \\
MaskGIT~\cite{chang2022maskgit} & 227M & 8 & 6.18 & 182.1 \\
MoVQ + NAR~\cite{zheng2022movq} & 389M & 12 & 7.22  & 130.1 \\ \midrule
SeQ-GAN + NAR & 229M & 12 & 4.99 & 189.1 \\ 
SeQ-GAN + NAR-L & 364M & 12 & \textbf{4.55} & \textbf{200.4} \\ \bottomrule
\end{tabular}
\vspace{-.14in}
\caption{\label{tab:condition}
FID and Inception Score (IS) comparison of conditional image generation on ImageNet~\cite{deng2009imagenet}.
}
\vspace{-.2in}
\end{table}

\subsection{Unconditional Image Generation}
We train AR and NAR transformers on top of SeQ-GAN for unconditional image generation on FFHQ~\cite{karras2019style} and LSUN~\cite{yu2015lsun} datasets. All models are trained for 500,000 iterations with the Adam optimizer, using a learning rate of 1e-4. Detailed hyperparameters are \textit{in the \secref{sec:setting}}.

From the results in \tabref{tab:uncondition},
previous state-of-the-art results are achieved by continuous diffusion model ADM~\cite{dhariwal2021diffusion} and StyleGAN2~\cite{karras2020analyzing}, while VQ-based generative models typically lag behind. 
Using SeQ-GAN as the VQ tokenizer enables our AR/NAR transformers with 171M parameters to surpass VIT-VQGAN~\cite{yu2021vector}, RQ-VAE~\cite{lee2022autoregressive}, and MoVQ~\cite{zheng2022movq}, despite having fewer parameters. Our method achieves comparable performance to ADM and StyleGAN2 on both FFHQ and LSUN datasets.

\subsection{Conditional Image Generation}
We train AR and NAR transformers with our SeQ-GAN tokenizer on 256$\times$256 ImageNet generation. The model is trained with a learning rate of 1e-4 for 300 epochs to enable direct comparison with VIT-VQGAN~\cite{yu2021vector} and MaskGIT~\cite{chang2022maskgit}. Further training settings can be found \textit{in the \secref{sec:setting}}.

\begin{table}[!tb]
\centering
\setlength\tabcolsep{3pt}
\begin{tabular}{l|cc|cc|cc}
\toprule
Model & Dim $\codebook$ & $K$ & Usage & rFID & AR & NAR \\ \midrule
VQGAN & 256 & 1024 & 43.5\% & 4.07 & 17.19 & 14.58 \\ \midrule
+ F\&N & 32 & 8192 & 99.7\% & \textbf{2.93} & 24.91 & - \\ 
+ K-means & 256 & 1024 & 100\% & 3.54 & \textbf{16.84} & 15.02 \\ 
+ $H(\hat{\mathcal{D}})$ & 256 & 1024 & 100\% & 3.45 & 16.97 & \textbf{13.26} \\ \bottomrule
\end{tabular}
\vspace{-.12in}
\caption{\label{tab:usage} Ablation study of codebook regularization compared to VQGAN~\cite{esser2021taming} baseline and VIT-VQGAN~\cite{yu2021vector} with factorized and L2-normed code (F\&N). $K$ denotes the codebook size.}
\vspace{-.15in}
\end{table}

\begin{table}[!tb]
\centering
\setlength\tabcolsep{3pt}
\begin{tabular}{l|cccccc|c|cc}
    Loss
    & \rotatebox{90}{relu1\_2}
    & \rotatebox{90}{relu2\_2}
    & \rotatebox{90}{relu3\_3}
    & \rotatebox{90}{relu4\_3}
    & \rotatebox{90}{relu5\_3}
    & \rotatebox{90}{logit}
    & \rotatebox{90}{rFID}
    & \rotatebox{90}{AR} 
    & \rotatebox{90}{NAR}
    \\
    \toprule
        $\mathcal{L}^{low}_{per}$ & \cmark & \cmark & \cmark & \cmark & \cmark & & 3.45 & 16.97 & 13.26 \\ \midrule
        A & \cmark & \cmark & \cmark & \cmark & \cmark & \cmark & 3.01 & 15.19 & 11.78 \\
        B & & \cmark & \cmark & \cmark & \cmark & \cmark & 2.93 & 14.47 & 11.58 \\
        C & & & \cmark & \cmark & \cmark & \cmark & 2.81 & 14.08 & 10.52 \\
        D & & & & \cmark & \cmark & \cmark & \textbf{2.62} & 13.34 & 9.56 \\
        $\mathcal{L}^{sem}_{per}$ & & & & & \cmark & \cmark & 2.77 & \textbf{12.07} & \textbf{8.84} \\
        E & & & & & & \cmark & 4.65 & 17.88 & 14.00 \\
    \bottomrule
    \end{tabular}
    \vspace{-.12in}
    \caption{\label{tab:semloss}
    Ablation of semantic-enhanced perceptual loss.}
    \vspace{-.25in}
\end{table}	

Results are summarized in \tabref{tab:condition}. 
Our SeQ-GAN+AR (364M, 256 sample steps) achieves FID of 6.25 and IS of 140.9, a remarkable improvement over VIT-VQGAN~\cite{yu2021vector} (714M, 256 sample steps), which
obtains 11.2 FID and 97.2 IS.
Compared to MaskGIT~\cite{chang2022maskgit}, which obtains 6.18 FID, our SeQ-GAN+NAR achieves a better 4.99 FID with a similar sampling step and model size.
Compared to MoVQ+NAR (389M, 12 sample steps), obtaining 7.22 FID and 130.1 IS, our SeQ-GAN+NAR-L (364M, 12 sample steps) achieves much better performance of 4.55 FID and 200.4 IS.

\subsection{Ablation}
\myPara{Codebook regularization.}
We ablate the strategy for increasing codebook usage in the baseline setting (one-phase training with $\mathcal{L}_{per}^{\alpha=0}$). As shown in \tabref{tab:usage}, although the factorized and L2-normed codebook in VIT-VQGAN can largely enhance the reconstruction fidelity, its large codebook size results in a suboptimal performance on the AR transformer. Moreover, optimizing the NAR transformer on a large codebook size is unstable. Compared to the offline K-means clustering used in previous codebook learning ~\cite{lancucki2020robust}, the entropy regularization used in our paper achieves a better reconstruction and generation performance.

\myPara{Design of semantic-enhanced perceptual loss.}
Our baseline, $\mathcal{L}^{low}_{per}$, utilizes all five layers to compute perceptual loss. As shown in \tabref{tab:semloss}, adding the logit feature improves both rFID and generation FID. Variant-D achieves the best rFID, while $\mathcal{L}^{sem}_{per}$ achieves the best generation FID. This demonstrates that reconstruction fidelity does not necessarily correlate with generation performance. Removing more shallow layers consistently improves generation quality, highlighting the importance of semantics when optimizing VQ tokenizers for generation quality. However, adopting the logit feature (variant-E) without the spatial feature results in significantly worse performance. While adjusting the balance between details and semantics by removing different perceptual layers is possible, it usually requires extensive parameter tuning to match the loss scale. Instead, we fix $\mathcal{L}^{low}_{per}$ and $\mathcal{L}^{sem}_{per}$ and simply tune the semantic ratio $\alpha$ in \eqref{eq:reweight} to achieve our goal.

\begin{figure}[!tb]
\centering
\subfloat[Effect of 2nd-phase tokenizer learning on generation quality.]{
\vspace{-.07in}
\includegraphics[width=0.98\linewidth]{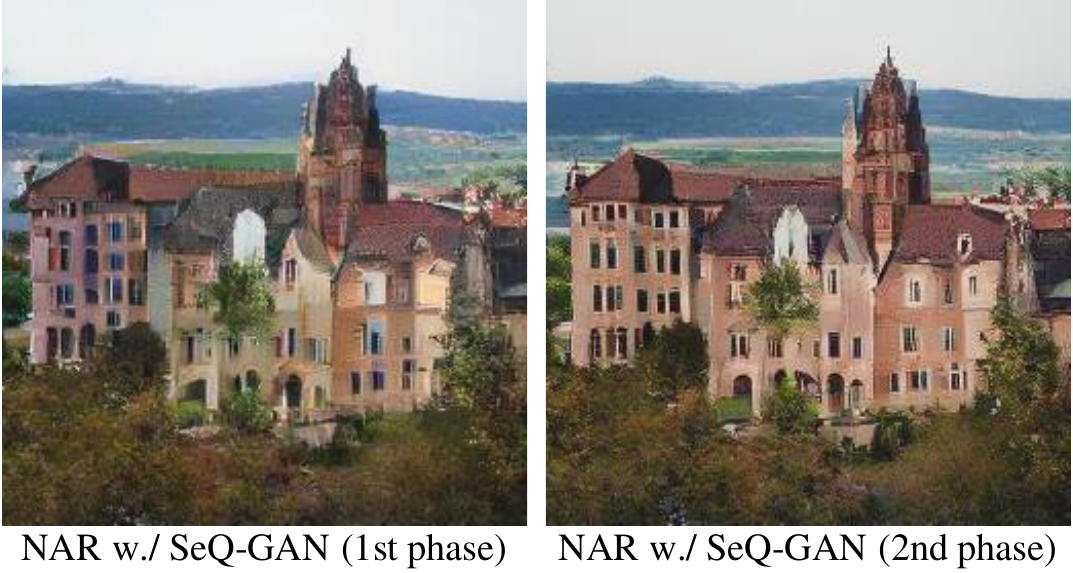}}\\
\subfloat[Effect of 2nd-phase tokenizer learning on generation FID.]{\vspace{-.07in}
\setlength\tabcolsep{4pt}
\begin{tabular}{l|c|ccccc}
Model                & SeQ-GAN & \rotatebox{90}{ImageNet} & \rotatebox{90}{FFHQ} & \rotatebox{90}{Church} & \rotatebox{90}{Cat}  & \rotatebox{90}{Bedroom} \\ \hline
\multirow{2}{*}{AR} & 1st phase & 7.83 & - & 3.49 & 4.73 & 2.15 \\ 
& 2nd phase & \textbf{7.55} & - & \textbf{2.45} & \textbf{3.61} & \textbf{1.44} \\ \hline
\multirow{2}{*}{NAR} & 1st phase & 5.31 & 3.89 & 3.41 & 5.22 & 2.88 \\  
& 2nd phase & \textbf{4.99} & \textbf{3.62} & \textbf{2.25} & \textbf{4.60} & \textbf{2.05} \\ \hline
\end{tabular}}
\vspace{-.1in}
\caption{\label{fig:phase2}
Ablation study on the impact of 2nd-phase tokenizer learning on generation.}
\vspace{-.2in}
\end{figure}

\myPara{Influence of the second phase tokenizer learning.}
SeQ-GAN is trained with semantic-enhanced perceptual loss $\mathcal{L}_{per}^{\alpha=1}$ in the first phase, which can result in some loss of color fidelity and high-frequency details. However, by finetuning the enhanced decoder in the second phase, those details can be preserved for the generation. As shown in \figref{fig:phase2}(a), the second phase learning can restore color distortion (\textit{e.g.}, windows). Furthermore, \figref{fig:phase2}(b) shows that second phase learning consistently improves generation FID. It's worth noting that joint learning the encoder/codebook and an enhanced decoder degrades the generation performance in our observation 1 (see \secref{sec:observation1}). Therefore, the decoder-only finetuning is an effective way to promote details preservation without degrading discrete latent space.

\section{Conclusion}
This work examines a fundamental question in VQ-based generative models, ``how the improved reconstruction of VQ
tokenizers affects the generation". To answer this question, we introduce a visualization pipeline to examine the influence of different tokenizers on AR transformers. Based on this pipeline, we find both semantic compression and details preservation should be considered in optimizing VQ tokenizers, in which previous works prioritize the latter. Based on this finding, we propose a simple solution SeQ-GAN, which achieves remarkable improvement over existing VQ-based generative models on image synthesis.

{\small
\bibliographystyle{ieee_fullname}
\bibliography{egbib}

\begin{thebibliography}{10}\itemsep=-1pt

\bibitem{austin2021structured}
Jacob Austin, Daniel~D Johnson, Jonathan Ho, Daniel Tarlow, and Rianne van~den
  Berg.
\newblock Structured denoising diffusion models in discrete state-spaces.
\newblock {\em NeurIPS}, 34:17981--17993, 2021.

\bibitem{bao2021beit}
Hangbo Bao, Li Dong, and Furu Wei.
\newblock Beit: Bert pre-training of image transformers.
\newblock {\em arXiv preprint arXiv:2106.08254}, 2021.

\bibitem{bengio2015scheduled}
Samy Bengio, Oriol Vinyals, Navdeep Jaitly, and Noam Shazeer.
\newblock Scheduled sampling for sequence prediction with recurrent neural
  networks.
\newblock {\em NeurIPS}, 28, 2015.

\bibitem{brock2018large}
Andrew Brock, Jeff Donahue, and Karen Simonyan.
\newblock Large scale gan training for high fidelity natural image synthesis.
\newblock {\em arXiv preprint arXiv:1809.11096}, 2018.

\bibitem{caron2020unsupervised}
Mathilde Caron, Ishan Misra, Julien Mairal, Priya Goyal, Piotr Bojanowski, and
  Armand Joulin.
\newblock Unsupervised learning of visual features by contrasting cluster
  assignments.
\newblock {\em NeurIPS}, 33:9912--9924, 2020.

\bibitem{chang2022maskgit}
Huiwen Chang, Han Zhang, Lu Jiang, Ce Liu, and William~T Freeman.
\newblock Maskgit: Masked generative image transformer.
\newblock In {\em CVPR}, pages 11315--11325, 2022.

\bibitem{chen2020generative}
Mark Chen, Alec Radford, Rewon Child, Jeffrey Wu, Heewoo Jun, David Luan, and
  Ilya Sutskever.
\newblock Generative pretraining from pixels.
\newblock In {\em ICML}, pages 1691--1703. PMLR, 2020.

\bibitem{chen2020improved}
Xinlei Chen, Haoqi Fan, Ross Girshick, and Kaiming He.
\newblock Improved baselines with momentum contrastive learning.
\newblock {\em arXiv preprint arXiv:2003.04297}, 2020.

\bibitem{chen2018pixelsnail}
Xi Chen, Nikhil Mishra, Mostafa Rohaninejad, and Pieter Abbeel.
\newblock Pixelsnail: An improved autoregressive generative model.
\newblock In {\em ICML}, pages 864--872. PMLR, 2018.

\bibitem{deng2009imagenet}
Jia Deng, Wei Dong, Richard Socher, Li-Jia Li, Kai Li, and Li Fei-Fei.
\newblock Imagenet: A large-scale hierarchical image database.
\newblock In {\em CVPR}, pages 248--255. Ieee, 2009.

\bibitem{dhariwal2021diffusion}
Prafulla Dhariwal and Alexander Nichol.
\newblock Diffusion models beat gans on image synthesis.
\newblock {\em NeurIPS}, 34:8780--8794, 2021.

\bibitem{ding2021cogview}
Ming Ding, Zhuoyi Yang, Wenyi Hong, Wendi Zheng, Chang Zhou, Da Yin, Junyang
  Lin, Xu Zou, Zhou Shao, Hongxia Yang, et~al.
\newblock Cogview: Mastering text-to-image generation via transformers.
\newblock {\em NeurIPS}, 34:19822--19835, 2021.

\bibitem{dong2021peco}
Xiaoyi Dong, Jianmin Bao, Ting Zhang, Dongdong Chen, Weiming Zhang, Lu Yuan,
  Dong Chen, Fang Wen, and Nenghai Yu.
\newblock Peco: Perceptual codebook for bert pre-training of vision
  transformers.
\newblock {\em arXiv preprint arXiv:2111.12710}, 2021.

\bibitem{esser2021imagebart}
Patrick Esser, Robin Rombach, Andreas Blattmann, and Bjorn Ommer.
\newblock Imagebart: Bidirectional context with multinomial diffusion for
  autoregressive image synthesis.
\newblock {\em NeurIPS}, 34:3518--3532, 2021.

\bibitem{esser2021taming}
Patrick Esser, Robin Rombach, and Bjorn Ommer.
\newblock Taming transformers for high-resolution image synthesis.
\newblock In {\em CVPR}, pages 12873--12883, 2021.

\bibitem{ge2022long}
Songwei Ge, Thomas Hayes, Harry Yang, Xi Yin, Guan Pang, David Jacobs, Jia-Bin
  Huang, and Devi Parikh.
\newblock Long video generation with time-agnostic vqgan and time-sensitive
  transformer.
\newblock {\em arXiv preprint arXiv:2204.03638}, 2022.

\bibitem{geirhos2018imagenet}
Robert Geirhos, Patricia Rubisch, Claudio Michaelis, Matthias Bethge, Felix~A
  Wichmann, and Wieland Brendel.
\newblock Imagenet-trained cnns are biased towards texture; increasing shape
  bias improves accuracy and robustness.
\newblock {\em arXiv preprint arXiv:1811.12231}, 2018.

\bibitem{goodfellow2020generative}
Ian Goodfellow, Jean Pouget-Abadie, Mehdi Mirza, Bing Xu, David Warde-Farley,
  Sherjil Ozair, Aaron Courville, and Yoshua Bengio.
\newblock Generative adversarial networks.
\newblock {\em Communications of the ACM}, 63(11):139--144, 2020.

\bibitem{gu2022vector}
Shuyang Gu, Dong Chen, Jianmin Bao, Fang Wen, Bo Zhang, Dongdong Chen, Lu Yuan,
  and Baining Guo.
\newblock Vector quantized diffusion model for text-to-image synthesis.
\newblock In {\em CVPR}, pages 10696--10706, 2022.

\bibitem{gu2022vqfr}
Yuchao Gu, Xintao Wang, Liangbin Xie, Chao Dong, Gen Li, Ying Shan, and
  Ming-Ming Cheng.
\newblock Vqfr: Blind face restoration with vector-quantized dictionary and
  parallel decoder.
\newblock {\em arXiv preprint arXiv:2205.06803}, 2022.

\bibitem{he2022masked}
Kaiming He, Xinlei Chen, Saining Xie, Yanghao Li, Piotr Doll{\'a}r, and Ross
  Girshick.
\newblock Masked autoencoders are scalable vision learners.
\newblock In {\em CVPR}, pages 16000--16009, 2022.

\bibitem{he2020momentum}
Kaiming He, Haoqi Fan, Yuxin Wu, Saining Xie, and Ross Girshick.
\newblock Momentum contrast for unsupervised visual representation learning.
\newblock In {\em CVPR}, pages 9729--9738, 2020.

\bibitem{ho2020denoising}
Jonathan Ho, Ajay Jain, and Pieter Abbeel.
\newblock Denoising diffusion probabilistic models.
\newblock {\em NeurIPS}, 33:6840--6851, 2020.

\bibitem{ho2022classifier}
Jonathan Ho and Tim Salimans.
\newblock Classifier-free diffusion guidance.
\newblock {\em arXiv preprint arXiv:2207.12598}, 2022.

\bibitem{holtzman2019curious}
Ari Holtzman, Jan Buys, Li Du, Maxwell Forbes, and Yejin Choi.
\newblock The curious case of neural text degeneration.
\newblock {\em arXiv preprint arXiv:1904.09751}, 2019.

\bibitem{hong2022cogvideo}
Wenyi Hong, Ming Ding, Wendi Zheng, Xinghan Liu, and Jie Tang.
\newblock Cogvideo: Large-scale pretraining for text-to-video generation via
  transformers.
\newblock {\em arXiv preprint arXiv:2205.15868}, 2022.

\bibitem{hoogeboom2021argmax}
Emiel Hoogeboom, Didrik Nielsen, Priyank Jaini, Patrick Forr{\'e}, and Max
  Welling.
\newblock Argmax flows and multinomial diffusion: Towards non-autoregressive
  language models.
\newblock 2021.

\bibitem{huang2017arbitrary}
Xun Huang and Serge Belongie.
\newblock Arbitrary style transfer in real-time with adaptive instance
  normalization.
\newblock In {\em ICCV}, pages 1501--1510, 2017.

\bibitem{johnson2016perceptual}
Justin Johnson, Alexandre Alahi, and Li Fei-Fei.
\newblock Perceptual losses for real-time style transfer and super-resolution.
\newblock In {\em ECCV}, pages 694--711. Springer, 2016.

\bibitem{karras2019style}
Tero Karras, Samuli Laine, and Timo Aila.
\newblock A style-based generator architecture for generative adversarial
  networks.
\newblock In {\em CVPR}, pages 4401--4410, 2019.

\bibitem{karras2020analyzing}
Tero Karras, Samuli Laine, Miika Aittala, Janne Hellsten, Jaakko Lehtinen, and
  Timo Aila.
\newblock Analyzing and improving the image quality of stylegan.
\newblock In {\em CVPR}, pages 8110--8119, 2020.

\bibitem{kingma2014adam}
Diederik~P Kingma and Jimmy Ba.
\newblock Adam: A method for stochastic optimization.
\newblock {\em arXiv preprint arXiv:1412.6980}, 2014.

\bibitem{kudo2018sentencepiece}
Taku Kudo and John Richardson.
\newblock Sentencepiece: A simple and language independent subword tokenizer
  and detokenizer for neural text processing.
\newblock {\em arXiv preprint arXiv:1808.06226}, 2018.

\bibitem{lancucki2020robust}
Adrian {\L}a{\'n}cucki, Jan Chorowski, Guillaume Sanchez, Ricard Marxer, Nanxin
  Chen, Hans~JGA Dolfing, Sameer Khurana, Tanel Alum{\"a}e, and Antoine
  Laurent.
\newblock Robust training of vector quantized bottleneck models.
\newblock In {\em IEEE IJCNN}, pages 1--7. IEEE, 2020.

\bibitem{lee2022autoregressive}
Doyup Lee, Chiheon Kim, Saehoon Kim, Minsu Cho, and Wook-Shin Han.
\newblock Autoregressive image generation using residual quantization.
\newblock In {\em CVPR}, pages 11523--11532, 2022.

\bibitem{li2022mc}
Xiaotong Li, Yixiao Ge, Kun Yi, Zixuan Hu, Ying Shan, and Ling-Yu Duan.
\newblock mc-beit: Multi-choice discretization for image bert pre-training.
\newblock {\em arXiv preprint arXiv:2203.15371}, 2022.

\bibitem{li2021contrastive}
Yunfan Li, Peng Hu, Zitao Liu, Dezhong Peng, Joey~Tianyi Zhou, and Xi Peng.
\newblock Contrastive clustering.
\newblock In {\em AAAI}, volume~35, pages 8547--8555, 2021.

\bibitem{nash2021generating}
Charlie Nash, Jacob Menick, Sander Dieleman, and Peter~W Battaglia.
\newblock Generating images with sparse representations.
\newblock {\em arXiv preprint arXiv:2103.03841}, 2021.

\bibitem{nichol2021improved}
Alexander~Quinn Nichol and Prafulla Dhariwal.
\newblock Improved denoising diffusion probabilistic models.
\newblock In {\em ICML}, pages 8162--8171. PMLR, 2021.

\bibitem{parmar2018image}
Niki Parmar, Ashish Vaswani, Jakob Uszkoreit, Lukasz Kaiser, Noam Shazeer,
  Alexander Ku, and Dustin Tran.
\newblock Image transformer.
\newblock In {\em ICML}, pages 4055--4064. PMLR, 2018.

\bibitem{peng2022beit}
Zhiliang Peng, Li Dong, Hangbo Bao, Qixiang Ye, and Furu Wei.
\newblock Beit v2: Masked image modeling with vector-quantized visual
  tokenizers.
\newblock {\em arXiv preprint arXiv:2208.06366}, 2022.

\bibitem{radford2021learning}
Alec Radford, Jong~Wook Kim, Chris Hallacy, Aditya Ramesh, Gabriel Goh,
  Sandhini Agarwal, Girish Sastry, Amanda Askell, Pamela Mishkin, Jack Clark,
  et~al.
\newblock Learning transferable visual models from natural language
  supervision.
\newblock In {\em ICML}, pages 8748--8763. PMLR, 2021.

\bibitem{radford2019language}
Alec Radford, Jeffrey Wu, Rewon Child, David Luan, Dario Amodei, Ilya
  Sutskever, et~al.
\newblock Language models are unsupervised multitask learners.
\newblock {\em OpenAI blog}, 1(8):9, 2019.

\bibitem{ramachandran2017searching}
Prajit Ramachandran, Barret Zoph, and Quoc~V Le.
\newblock Searching for activation functions.
\newblock {\em arXiv preprint arXiv:1710.05941}, 2017.

\bibitem{ramesh2021zero}
Aditya Ramesh, Mikhail Pavlov, Gabriel Goh, Scott Gray, Chelsea Voss, Alec
  Radford, Mark Chen, and Ilya Sutskever.
\newblock Zero-shot text-to-image generation.
\newblock In {\em ICML}, pages 8821--8831. PMLR, 2021.

\bibitem{razavi2019generating}
Ali Razavi, Aaron Van~den Oord, and Oriol Vinyals.
\newblock Generating diverse high-fidelity images with vq-vae-2.
\newblock {\em NeurIPS}, 32, 2019.

\bibitem{rombach2022high}
Robin Rombach, Andreas Blattmann, Dominik Lorenz, Patrick Esser, and Bj{\"o}rn
  Ommer.
\newblock High-resolution image synthesis with latent diffusion models.
\newblock In {\em CVPR}, pages 10684--10695, 2022.

\bibitem{sennrich2015neural}
Rico Sennrich, Barry Haddow, and Alexandra Birch.
\newblock Neural machine translation of rare words with subword units.
\newblock {\em arXiv preprint arXiv:1508.07909}, 2015.

\bibitem{simonyan2014very}
Karen Simonyan and Andrew Zisserman.
\newblock Very deep convolutional networks for large-scale image recognition.
\newblock {\em arXiv preprint arXiv:1409.1556}, 2014.

\bibitem{tang2022improved}
Zhicong Tang, Shuyang Gu, Jianmin Bao, Dong Chen, and Fang Wen.
\newblock Improved vector quantized diffusion models.
\newblock {\em arXiv preprint arXiv:2205.16007}, 2022.

\bibitem{van2016conditional}
Aaron Van~den Oord, Nal Kalchbrenner, Lasse Espeholt, Oriol Vinyals, Alex
  Graves, et~al.
\newblock Conditional image generation with pixelcnn decoders.
\newblock {\em NeurIPS}, 29, 2016.

\bibitem{van2017neural}
Aaron Van Den~Oord, Oriol Vinyals, et~al.
\newblock Neural discrete representation learning.
\newblock {\em NeurIPS}, 30, 2017.

\bibitem{wang2022restoreformer}
Zhouxia Wang, Jiawei Zhang, Runjian Chen, Wenping Wang, and Ping Luo.
\newblock Restoreformer: High-quality blind face restoration from undegraded
  key-value pairs.
\newblock In {\em CVPR}, pages 17512--17521, 2022.

\bibitem{williams1989learning}
Ronald~J Williams and David Zipser.
\newblock A learning algorithm for continually running fully recurrent neural
  networks.
\newblock {\em Neural computation}, 1(2):270--280, 1989.

\bibitem{wu2018group}
Yuxin Wu and Kaiming He.
\newblock Group normalization.
\newblock In {\em ECCV}, pages 3--19, 2018.

\bibitem{xie2022simmim}
Zhenda Xie, Zheng Zhang, Yue Cao, Yutong Lin, Jianmin Bao, Zhuliang Yao, Qi
  Dai, and Han Hu.
\newblock Simmim: A simple framework for masked image modeling.
\newblock In {\em CVPR}, pages 9653--9663, 2022.

\bibitem{yan2021videogpt}
Wilson Yan, Yunzhi Zhang, Pieter Abbeel, and Aravind Srinivas.
\newblock Videogpt: Video generation using vq-vae and transformers.
\newblock {\em arXiv preprint arXiv:2104.10157}, 2021.

\bibitem{yu2015lsun}
Fisher Yu, Ari Seff, Yinda Zhang, Shuran Song, Thomas Funkhouser, and Jianxiong
  Xiao.
\newblock Lsun: Construction of a large-scale image dataset using deep learning
  with humans in the loop.
\newblock {\em arXiv preprint arXiv:1506.03365}, 2015.

\bibitem{yu2021vector}
Jiahui Yu, Xin Li, Jing~Yu Koh, Han Zhang, Ruoming Pang, James Qin, Alexander
  Ku, Yuanzhong Xu, Jason Baldridge, and Yonghui Wu.
\newblock Vector-quantized image modeling with improved vqgan.
\newblock {\em arXiv preprint arXiv:2110.04627}, 2021.

\bibitem{yu2022scaling}
Jiahui Yu, Yuanzhong Xu, Jing~Yu Koh, Thang Luong, Gunjan Baid, Zirui Wang,
  Vijay Vasudevan, Alexander Ku, Yinfei Yang, Burcu~Karagol Ayan, et~al.
\newblock Scaling autoregressive models for content-rich text-to-image
  generation.
\newblock {\em arXiv preprint arXiv:2206.10789}, 2022.

\bibitem{zhang2018unreasonable}
Richard Zhang, Phillip Isola, Alexei~A Efros, Eli Shechtman, and Oliver Wang.
\newblock The unreasonable effectiveness of deep features as a perceptual
  metric.
\newblock In {\em CVPR}, pages 586--595, 2018.

\bibitem{zhao2021improved}
Long Zhao, Zizhao Zhang, Ting Chen, Dimitris Metaxas, and Han Zhang.
\newblock Improved transformer for high-resolution gans.
\newblock {\em NeurIPS}, 34:18367--18380, 2021.

\bibitem{zheng2022movq}
Chuanxia Zheng, Long~Tung Vuong, Jianfei Cai, and Dinh Phung.
\newblock Movq: Modulating quantized vectors for high-fidelity image
  generation.
\newblock {\em arXiv preprint arXiv:2209.09002}, 2022.

\bibitem{zhou2021ibot}
Jinghao Zhou, Chen Wei, Huiyu Wang, Wei Shen, Cihang Xie, Alan Yuille, and Tao
  Kong.
\newblock ibot: Image bert pre-training with online tokenizer.
\newblock {\em arXiv preprint arXiv:2111.07832}, 2021.

\bibitem{zhou2022towards}
Shangchen Zhou, Kelvin~CK Chan, Chongyi Li, and Chen~Change Loy.
\newblock Towards robust blind face restoration with codebook lookup
  transformer.
\newblock {\em arXiv preprint arXiv:2206.11253}, 2022.

\end{thebibliography}
}

\clearpage
\onecolumn
\section{Appendix}

In this section, we first present detailed experimental settings in \secref{sec:setting}. Next, in \secref{sec:vis_observation},
we offer additional visualization and analysis to further our understanding of the observations. We then present more qualitative results of our method in  \secref{sec:qualitative_results}. Finally, we discuss the limitations of our approach and potential directions for future work in \secref{sec:limitation}.

\begin{table*}[h]
    \centering
    \begin{tabular}{l | c | c }
    \toprule
        Input size & Encoder & Decoder\\\midrule
    \multirow{1}*{$f1:$ 256$\times$256} & \makecell[c]{$
        \begin{array}{c}
        \text{Conv, c-128}\\ \hline
        \left\{
        \text{Residual Block, 128-c}
        \right\}\times 2\\ \hline
        \text{Downsample Block, 128-c}
        \end{array}
        $} & \makecell[c]{$
        \begin{array}{c}
        \text{GN-Swish-Conv, c-3}\\ \hline
        \left\{\text{Residual Block, 128-c}\right\} \times 2 + \color{cyan}{\text{B\&D Attn}}
        \end{array}$}
         \\\midrule
         \multirow{1}*{$f2:$ 128$\times$128} & \makecell[c]{$
        \begin{array}{c}
        \left\{
        \text{Residual Block, 128-c}
        \right\}\times 2\\ \hline
        \text{Downsample Block, 256-c}
        \end{array}
        $} & \makecell[c]{$
        \begin{array}{c}
        \text{Upsample Block, c-128}\\\hline
        \left\{\text{Residual Block, 256-c}\right\} \times 2 + \color{cyan}{\text{B\&D Attn}}
        \end{array}$}
         \\\midrule
         \multirow{1}*{$f3:$ 64$\times$64} & \makecell[c]{$
        \begin{array}{c}
        \left\{
        \text{Residual Block, 256-c}
        \right\}\times 2\\ \hline
        \text{Downsample Block, 256-c}
        \end{array}
        $} & \makecell[c]{$
        \begin{array}{c}
        \text{Upsample Block, c-256}\\\hline
        \left\{\text{Residual Block, 256-c}\right\} \times 2 + \color{cyan}{\text{B\&D Attn}}
        \end{array}$}
         \\\midrule
         \multirow{1}*{$f4:$ 32$\times$32} & \makecell[c]{$
        \begin{array}{c}
        \left\{
        \text{Residual Block, 256-c}
        \right\}\times 2\\ \hline
        \text{Downsample Block, 512-c}
        \end{array}
        $} & \makecell[c]{$
        \begin{array}{c}
        \text{Upsample Block, c-256}\\\hline
        \left\{\text{Residual Block, 256-c}\right\} \times 2 + \color{cyan}{\text{B\&D Attn}}
        \end{array}$}
         \\\midrule
         \multirow{1}*{$f5:$ 16$\times$16} & \makecell[c]{$
        \begin{array}{c}
        \left\{
        \text{Residual Block, 512-c}
        \right\}\times 4\\ \hline
        \text{GN-Swish-Conv, 256-c}
        \end{array}
        $} & \makecell[c]{$
        \begin{array}{c}
        \text{Upsample Block, c-256}\\\hline
        \left\{\text{Residual Block, 512-c}\right\} \times 4 + \color{cyan}{\text{B\&D Attn}}\\ \hline
        \text{Conv, c-512}
        \end{array}$}
         \\
    \bottomrule
    \end{tabular}    \caption{\label{tab:arch_detail}Architecture of SeQ-GAN for 1st phase and \textcolor{cyan}{2nd phase} tokenizer learning.
    The residual block consists of GN~\cite{wu2018group}-Swish~\cite{ramachandran2017searching}-Conv-GN-Swish-Conv. B\&D Attn: interleaved block regional and dilated attention~\cite{zhao2021improved}; c: channels; f: compression ratio.}
    \vspace{-.2in}
\end{table*}

\begin{table*}[h]
\centering
\begin{tabular}{l|ccccccc}
\toprule
Model          & \#Params & \#Blocks & \#Heads & Model Dim & Hidden Dim & Dropout & \#Tokens \\ \midrule
AR/NAR & 172M     & 24       & 16      & 768       & 3072       & 0.1     & 256      \\
AR/NAR (Large) & 305M     & 24       & 16      & 1024      & 4096       & 0.1     & 256      \\ \bottomrule
\end{tabular} \caption{\label{tab:transformer_detail}Architecture of autoregressive (AR) and non-autoregressive (NAR) transformers. Both transformers share the same architecture, except for the causal attention used in the AR transformer.}
 \vspace{-.2in}
\end{table*}

\subsection{Experimental Settings}
\label{sec:setting}
\noindent\textbf{Tokenizer learning.} As shown in \tabref{tab:arch_detail}, SeQ-GAN's architecture is based on VQGAN~\cite{esser2021taming}. However, we modified the architecture in the first learning phase by removing the attention and constructing a convolution-only VQGAN. In the second learning phase, we enhanced the decoder with block regional and dilated attention (B$\&$D Attn) to make attention suitable for high-resolution feature maps. SeQ-GAN has a total of 54.5M and 57.9M parameters for the first and second learning phases, respectively. We use the style-based discriminator~\cite{karras2020analyzing} for training SeQ-GAN, as suggested in VIT-VQGAN~\cite{yu2021vector}. The hyperparameters used for training SeQ-GAN are summarized in \tabref{tab:setting_codebook}.

\noindent\textbf{Generative transformer training.}
The autoregressive (AR) and non-autoregressive (NAR) transformers share the same architecture, except that the AR transformer adopts causal attention. As detailed in \tabref{tab:transformer_detail}, the AR/NAR transformers and their large variant have 172M and 305M parameters, respectively. We train both types of generative transformer using the hyperparameters listed in  \tabref{tab:setting_transformer}. During sampling, we adopt the basic sampling techniques from VQGAN~\cite{esser2021taming} (\textit{i.e.}, top-$p$ sampling~\cite{holtzman2019curious}) and MaskGIT~\cite{chang2022maskgit} (\textit{i.e.}, adjusting sample temperature), while excluding the classifier-free guidance~\cite{ho2022classifier} and rejection sampling~\cite{ramesh2021zero} for simplicity.

\noindent\textbf{Detailed settings for observation and ablation experiments.}
Our observation (see Sec. 3.3 in the manuscript) and ablation experiments (see Sec. 4.4 in the manuscript) are conducted on the ImageNet~\cite{deng2009imagenet} dataset. 
We mostly follow the same configurations as the benchmark experiments listed in \tabref{tab:setting_codebook}, with the exception that we use a batch size of 64 for SeQ-GAN learning.
Based on each VQ tokenizer, we train the generative transformer on ImageNet with a batch size of 64 for 500,000 iterations, while keeping other settings the same as in \tabref{tab:setting_transformer}.

For Observation 1 (see Sec. 3.3.1), we evaluate different VQ tokenizers on various transformer configurations:
1) Different parameter sizes, including AR with 172M parameters and AR-Large with 305M parameters.
2) Different types of transformers, including both autoregressive and non-autoregressive transformers with 172M parameters.
3) Different training iterations, including AR-Large and AR-Large-2$\times$, where we add an extra 500,000 iterations to the AR-Large model to investigate whether longer training eliminates the difference in VQ tokenizer.

\begin{table*}[h]
\centering
\begin{tabular}{l|c|c|c|c|c}
\toprule  & ImageNet & FFHQ & Cat & Bedroom & Church \\ \hline\hline
\multicolumn{6}{c}{Dataset Statistics} \\ \hline
Training Set & 1,281,167 & 60,000 & 1,657,266 & 3,033,042 & 126,227 \\ 
Validation Set & 50,000 & 10,000 & - & - & - \\ \hline\hline
\multicolumn{6}{c}{1st Phase of Tokenizer Learning} \\ \hline
Batch Size & 256 & 64 & 64 & 64 & 32 \\
Iterations & 500,000 & 300,000 & 26,000 & 48,000 & 4,000 \\ \cline{4-6}
Epochs & 100 & 320 & \multicolumn{3}{c}{1} \\\cline{2-3}
Learning Rate & \multicolumn{2}{c|}{1e-4} & \multicolumn{3}{c}{5e-5} \\
LR Decay & \multicolumn{2}{c|}{Cosine ($end\_lr$=5e-5)} & \multicolumn{3}{c}{-} \\ \cline{2-6}
Optimizer & \multicolumn{5}{c}{Adam ($\beta_1$=0.9, $\beta_2$=0.99) }\\
\hline\hline
\multicolumn{6}{c}{2nd Phase of Tokenizer Learning} \\ \hline
Batch Size & 128 & 32 & 64 & 64 & 32  \\ \cline{2-6} 
Iterations & \multicolumn{5}{c}{200,000}\\
Learning Rate & \multicolumn{5}{c}{5e-5}\\
Optimizer & \multicolumn{5}{c}{Adam ($\beta_1$=0.5, $\beta_2$=0.9) }\\
\bottomrule
\end{tabular}
\caption{\label{tab:setting_codebook} Experimental setting of training SeQ-GAN on ImageNet~\cite{deng2009imagenet}, FFHQ~\cite{karras2019style}, and LSUN~\cite{yu2015lsun}-\{Cat, Bedroom, Church\}.}
\vspace{-.15in}
\end{table*}

\begin{table*}[h]
\centering
\begin{tabular}{l|c|c|c|c|c}
\toprule  & ImageNet & FFHQ & Cat & Bedroom & Church \\ \hline\hline
\multicolumn{6}{c}{Dataset Statistics} \\ \hline
Training Set & 1,281,167 & 60,000 & 1,657,266 & 3,033,042 & 126,227 \\ 
Validation Set & 50,000 & 10,000 & - & - & - \\ \hline\hline
\multicolumn{6}{c}{Autoregressive Transformer (AR)} \\ \hline
Batch Size & 256 & 32 & 256 & 256 & 64 \\ \cline{3-6}
Iterations & 1,500,000 & \multicolumn{4}{c}{500,000} \\ \cline{2-6}
Optimizer & \multicolumn{5}{c}{AdamW ($\beta_1$=0.9, $\beta_2$=0.96, weight\_decay=1e-2) }\\
Learning Rate & \multicolumn{5}{c}{1e-4}\\
LR Decay & \multicolumn{5}{c}{Exponential ($end\_lr$=5e-6, $start\_iter=80,000$)} \\ \cline{2-6}
Top-$p$ Sampling & 0.92 & \multicolumn{4}{c}{0.98}\\
\hline\hline
\multicolumn{6}{c}{Non-Autoregressive Transformer (NAR)} \\ \hline
Batch Size & 256 & 32 & 256 & 256 & 64 \\ \cline{3-6}
Iterations & 1,500,000 & \multicolumn{4}{c}{500,000} \\ \cline{2-6}
Optimizer & \multicolumn{5}{c}{AdamW ($\beta_1$=0.9, $\beta_2$=0.96, weight\_decay=1e-2) }\\
Learning Rate & \multicolumn{5}{c}{1e-4}\\
LR Decay & \multicolumn{5}{c}{Linear ($end\_lr$=0, $start\_iter=50,000$)} \\ \cline{2-6}
Sampling Temperature & 0.45 & \multicolumn{4}{c}{0.65}\\
\bottomrule
\end{tabular}
\caption{\label{tab:setting_transformer} Experimental setting of training generative transformers on ImageNet~\cite{deng2009imagenet}, FFHQ~\cite{karras2019style}, and LSUN~\cite{yu2015lsun}-\{Cat, Bedroom, Church\}.}
\end{table*}

\begin{figure*}[!tb]
    \centering
    \includegraphics[width=0.9\linewidth]{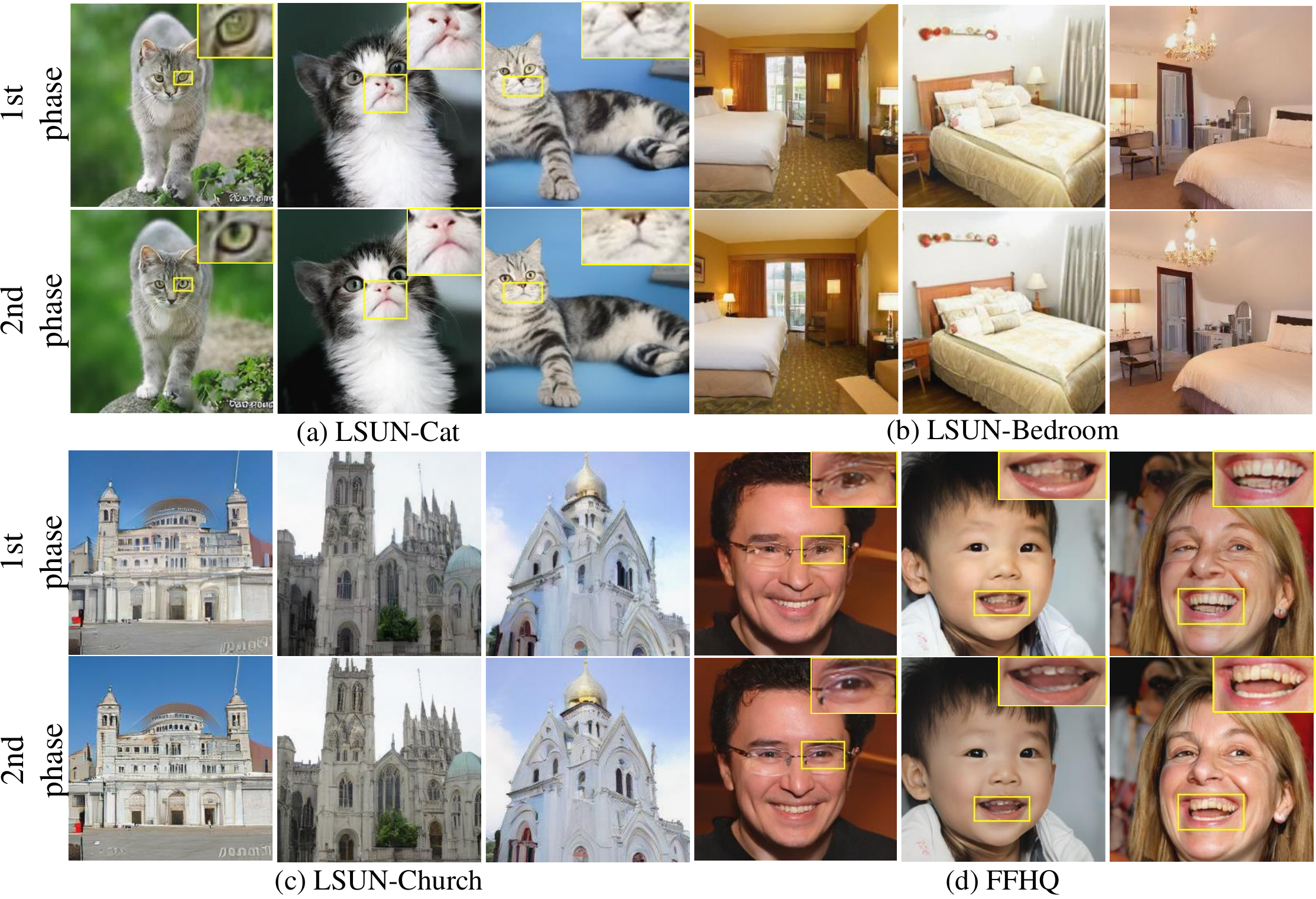}    \caption{\label{fig:2ndphase_vis}Influence of the 2nd phase tokenizer learning on generation results. (\textbf{Zoom in for best view.})}
    \vspace{-.2in}
\end{figure*}

\subsection{More Visualization and Analysis of the Observations}
\label{sec:vis_observation}
In this section, we present additional visualizations to support our observations and proposed solutions.

\begin{figure*}[h]
    \centering
\includegraphics[width=0.8\linewidth]{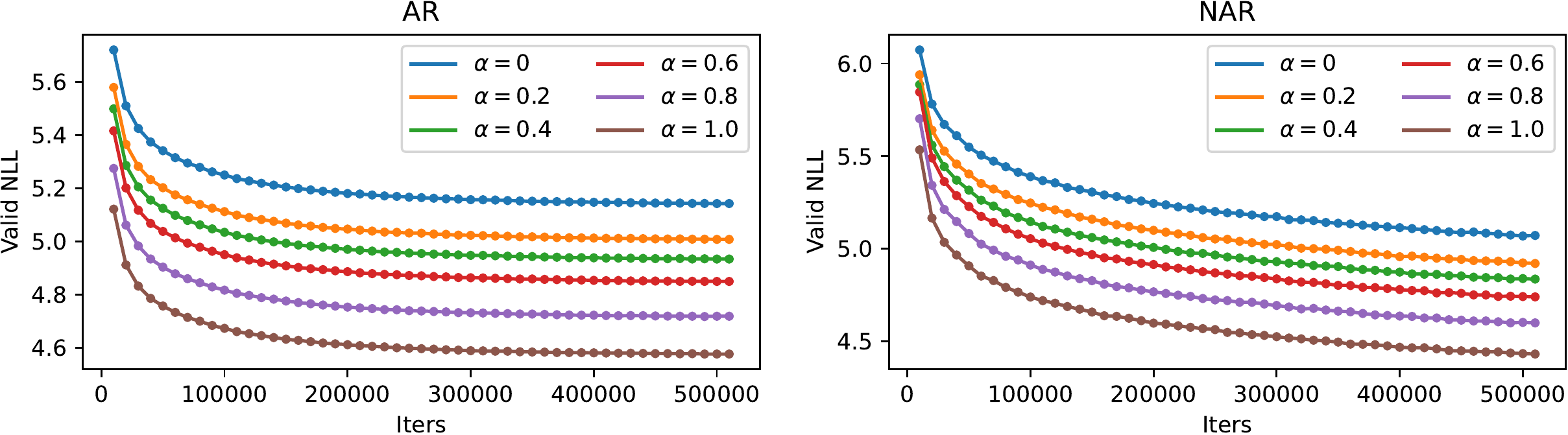}
\vspace{-.12in}
\caption{\label{fig:semratio_loss}Validation loss curves of generative transformers training on ImageNet. Generative transformers are built upon the SeQ-GAN tokenizers with different semantic ratios ($\alpha$).}
    \vspace{-.15in}
\end{figure*}

First, we train SeQ-GAN with varying semantic ratios $\alpha$ and plot the validation loss curve for each corresponding generative transformer training in \figref{fig:semratio_loss}. Our results show that a larger semantic ratio $\alpha$ results in lower validation loss, indicating that generative transformers are better able to model the discrete space constructed by VQ tokenizers when more semantics are incorporated.

\begin{figure*}[h]
    \centering
    \includegraphics[width=0.9\linewidth]{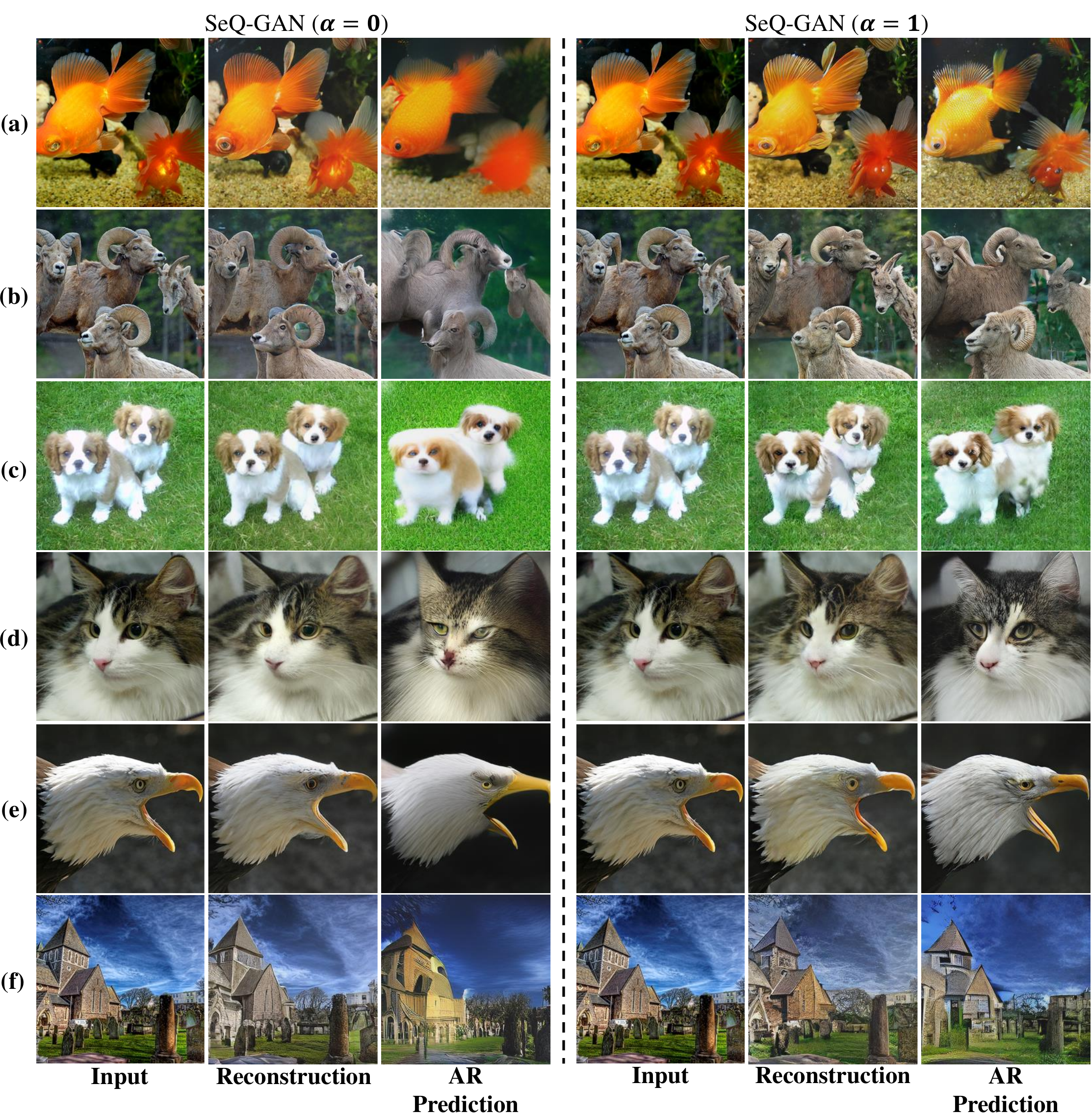}    \caption{\label{fig:arcom_vis}Visual comparison of the influence of SeQ-GAN with different semantic ratios ($\alpha\in{0,1}$) on image reconstruction and AR prediction. Compared to SeQ-GAN ($\alpha$=0), the AR transformer built on SeQ-GAN ($\alpha$=1) better models each instance (\textit{e.g.}, Row (a-c)), semantic features (\textit{e.g.}, the cat's face in Row (d) and the eagle's beak in Row (e)), and structure (\textit{e.g.}, the peaked roof in Row (f)).}
    \vspace{-.15in}
\end{figure*}

Next, we employ our proposed visualization pipeline to examine the reconstruction and AR prediction using SeQ-GAN with two different semantic ratios ($\alpha\in{0,1}$). \figref{fig:arcom_vis} demonstrates that the generative transformer trained on SeQ-GAN ($\alpha$=1) is better able to model each instance (\textit{e.g.}, Row (a-c)), the semantic features (\textit{e.g.}, the cat's face in Row (d) and the eagle's beak in Row (e)), and the structure (\textit{e.g.}, the peaked roof in Row (f)). Note that compared to the SeQ-GAN ($\alpha$=0) in \figref{fig:arcom_vis}, the reconstruction of the SeQ-GAN ($\alpha$=1) loses some color fidelity and high-frequency details, leading to similar problems of lost details and spatial distortion in the generation results (see \figref{fig:2ndphase_vis} (1st phase)).

Finally, to address the issue of lost details and spatial distortion resulting from removing shallow layers in $\mathcal{L}_{per}^{\alpha=1}$ during the first phase of tokenizer training, we use a two-phase tokenizer learning approach in our SeQ-GAN.
In the second phase, we finetune an enhanced decoder to restore the lost details. 
To demonstrate the effectiveness of our two-phase tokenizer learning on generation quality, we decode the transformer-sampled indices to image space using the decoder from both SeQ-GAN (1st phase) and SeQ-GAN (2nd phase), and present the generation results in \figref{fig:2ndphase_vis}. Our visualization clearly shows that SeQ-GAN (2nd phase) preserves more details and enhances generation quality compared to SeQ-GAN (1st phase).

\newpage

\subsection{More Qualitative Results}
\label{sec:qualitative_results}
\begin{figure*}[h]
    \centering
    \includegraphics[width=\linewidth]{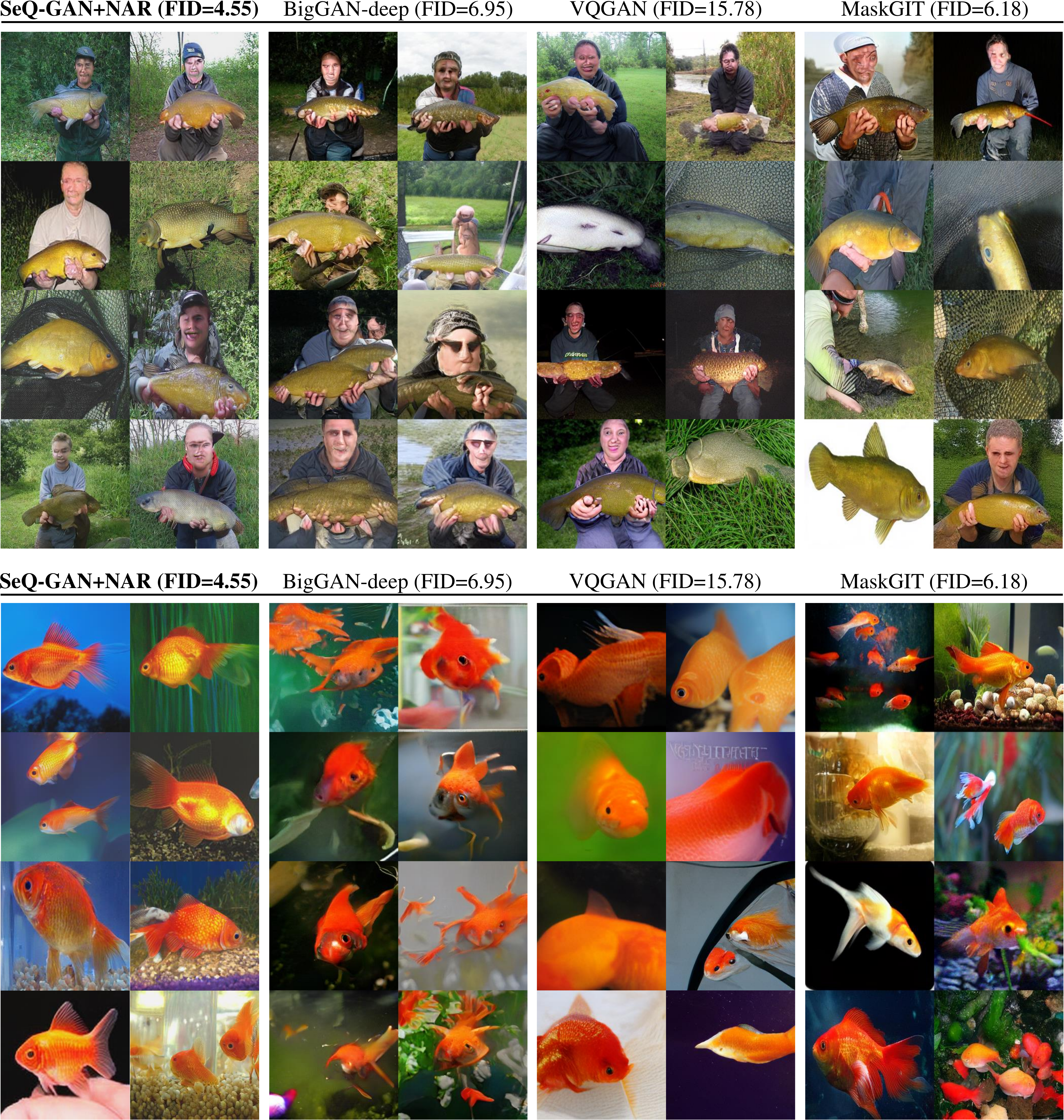}    \caption{\label{fig:imagenet_cls0_1}Qualitative comparison with BigGAN~\cite{brock2018large}, VQGAN~\cite{esser2021taming} and MaskGIT~\cite{chang2022maskgit} on the class 0 (tench) and class 1 (glodfish) of ImageNet~\cite{deng2009imagenet}. }
\end{figure*}

\begin{figure*}[h]
    \centering
    \includegraphics[width=\linewidth]{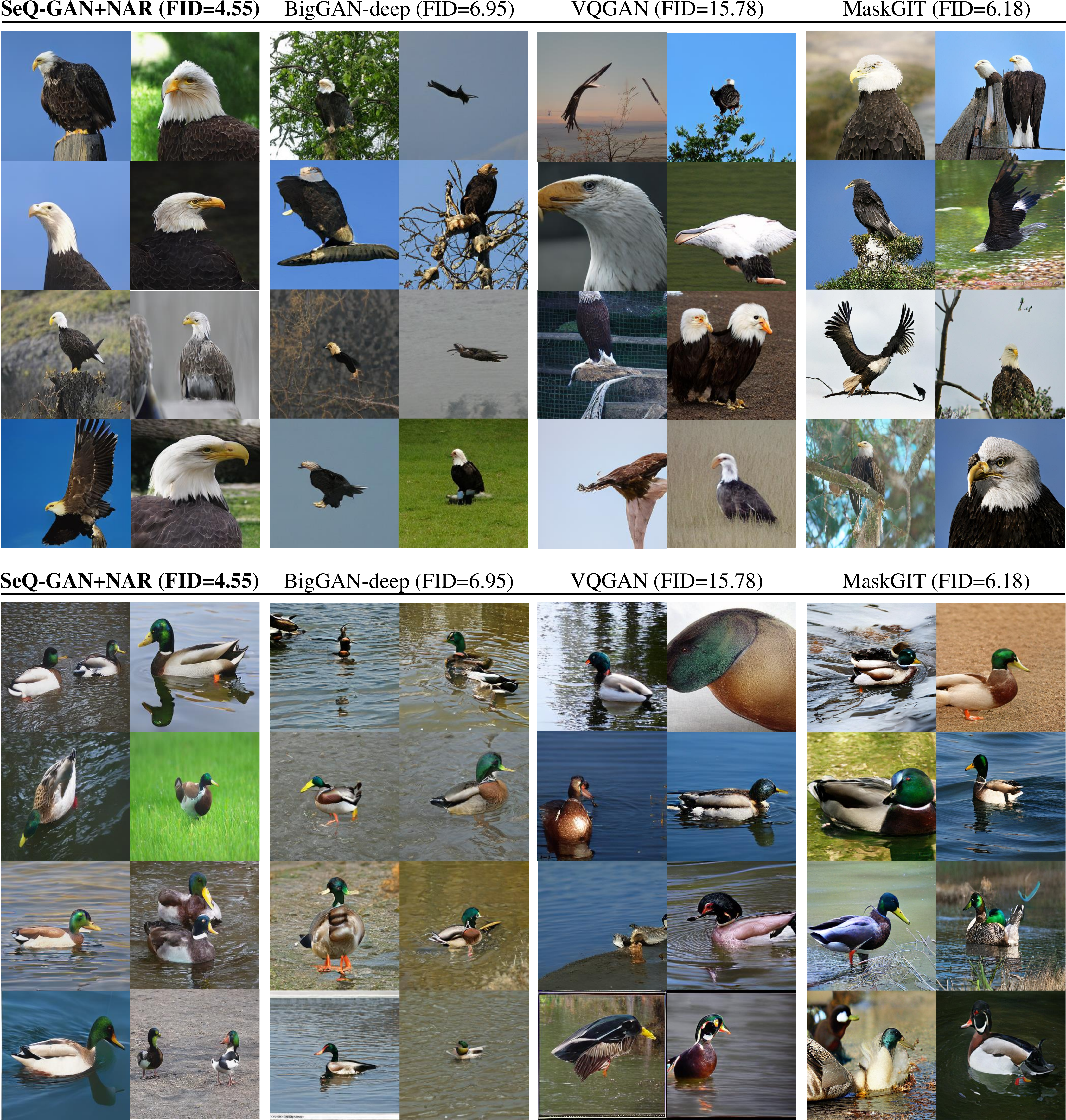}    \caption{\label{fig:imagenet_cls22_97}Qualitative comparison with BigGAN~\cite{brock2018large}, VQGAN~\cite{esser2021taming} and MaskGIT~\cite{chang2022maskgit} on the class 22 (bald eagle) and class 97 (drake) of ImageNet~\cite{deng2009imagenet}.}
\end{figure*}

\begin{figure*}[h]
    \centering
    \includegraphics[width=\linewidth]{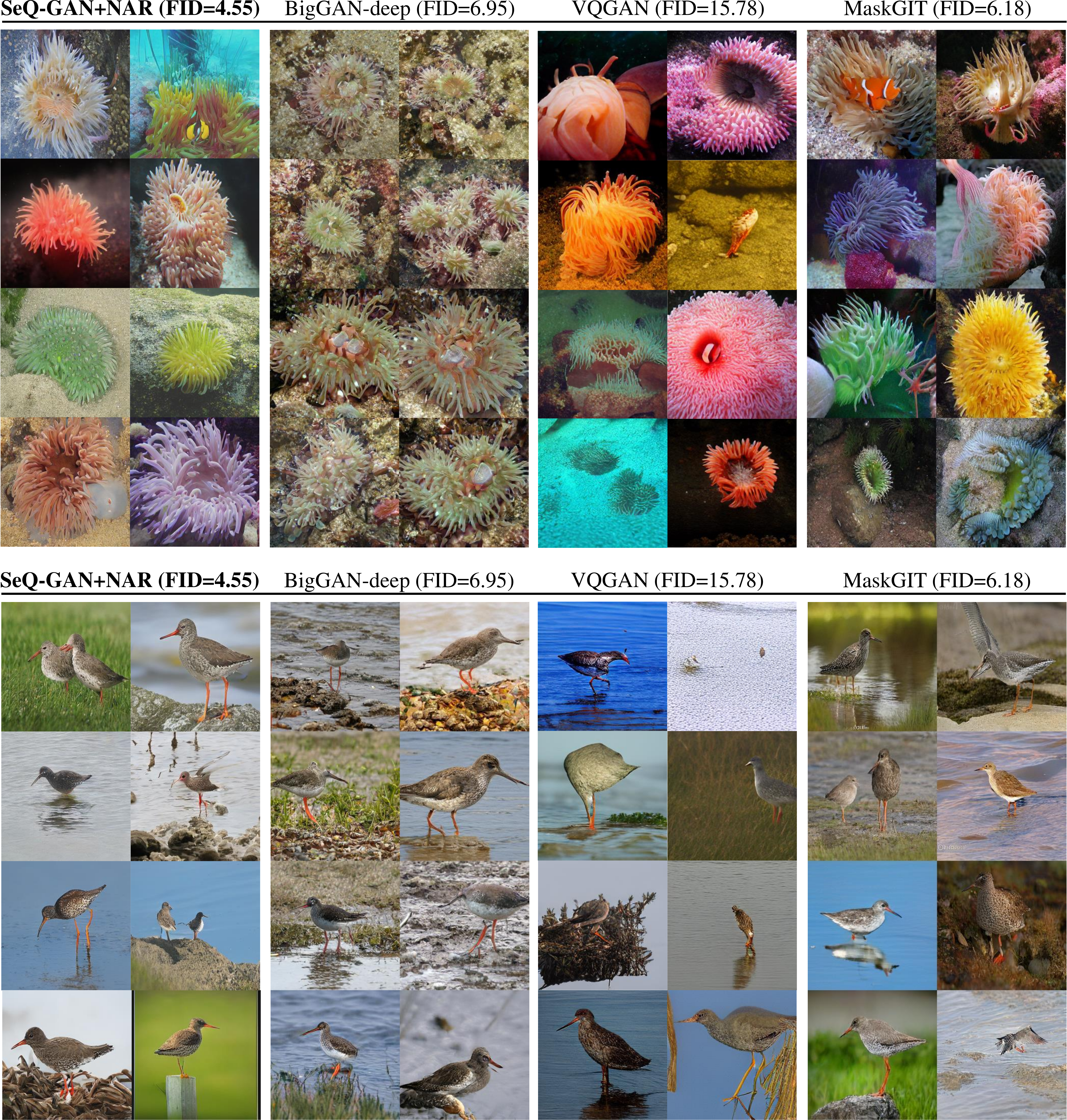}    \caption{\label{fig:imagenet_cls108_141}Qualitative comparison with BigGAN~\cite{brock2018large}, VQGAN~\cite{esser2021taming} and MaskGIT~\cite{chang2022maskgit} on the class 108 (sea anemone) and class 141 (redshank) of ImageNet~\cite{deng2009imagenet}.}
\end{figure*}

\begin{figure*}[h]
    \centering
    \includegraphics[width=\linewidth]{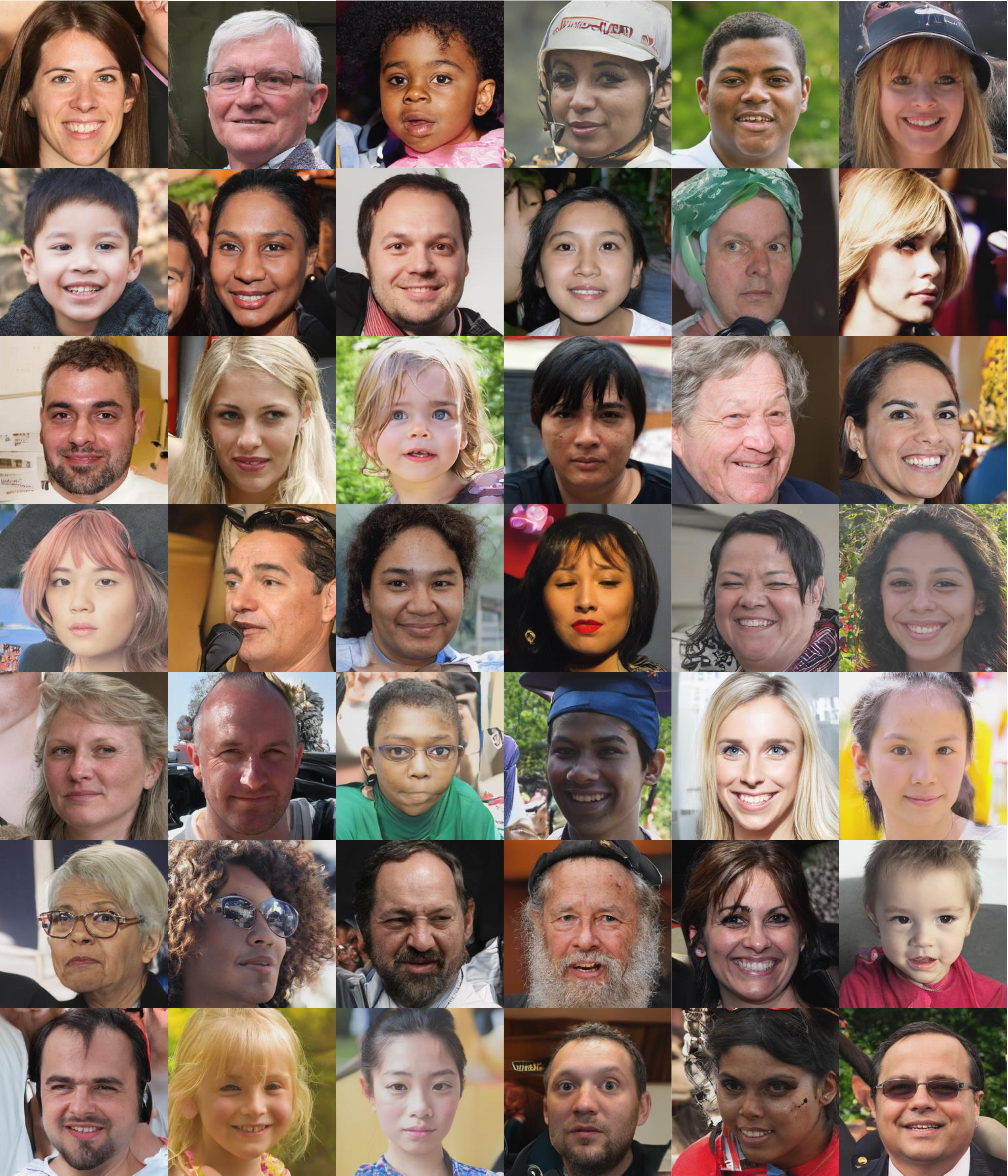}    \caption{\label{fig:uncurate_ffhq}Uncurated set of samples of SeQ-GAN+NAR on 256$\times$256 FFHQ.}
\end{figure*}

\begin{figure*}[h]
    \centering
    \includegraphics[width=\linewidth]{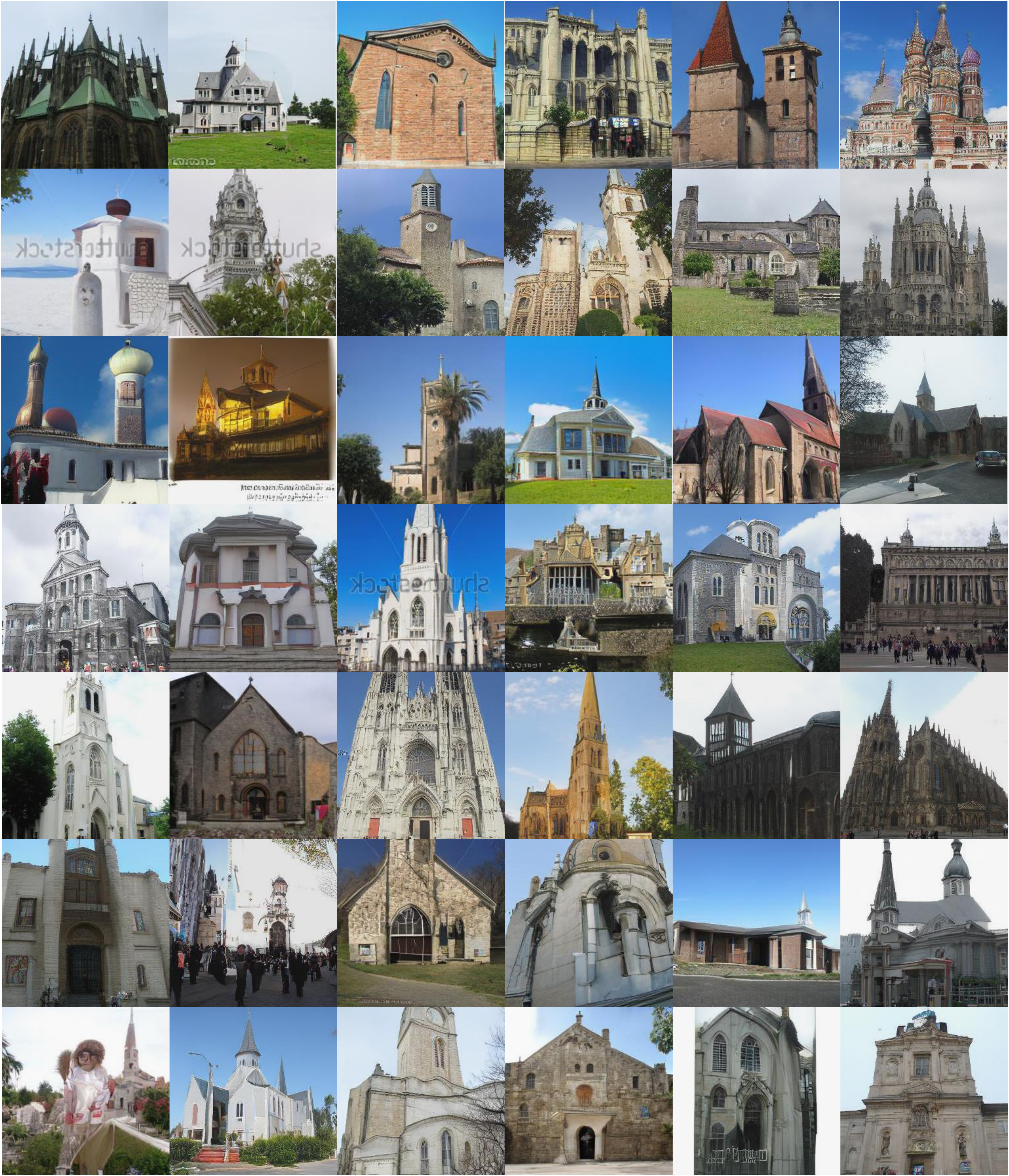}    \caption{\label{fig:uncurate_church}Uncurated set of samples of on 256$\times$256 LSUN church.}
\end{figure*}

\begin{figure*}[h]
    \centering
    \includegraphics[width=\linewidth]{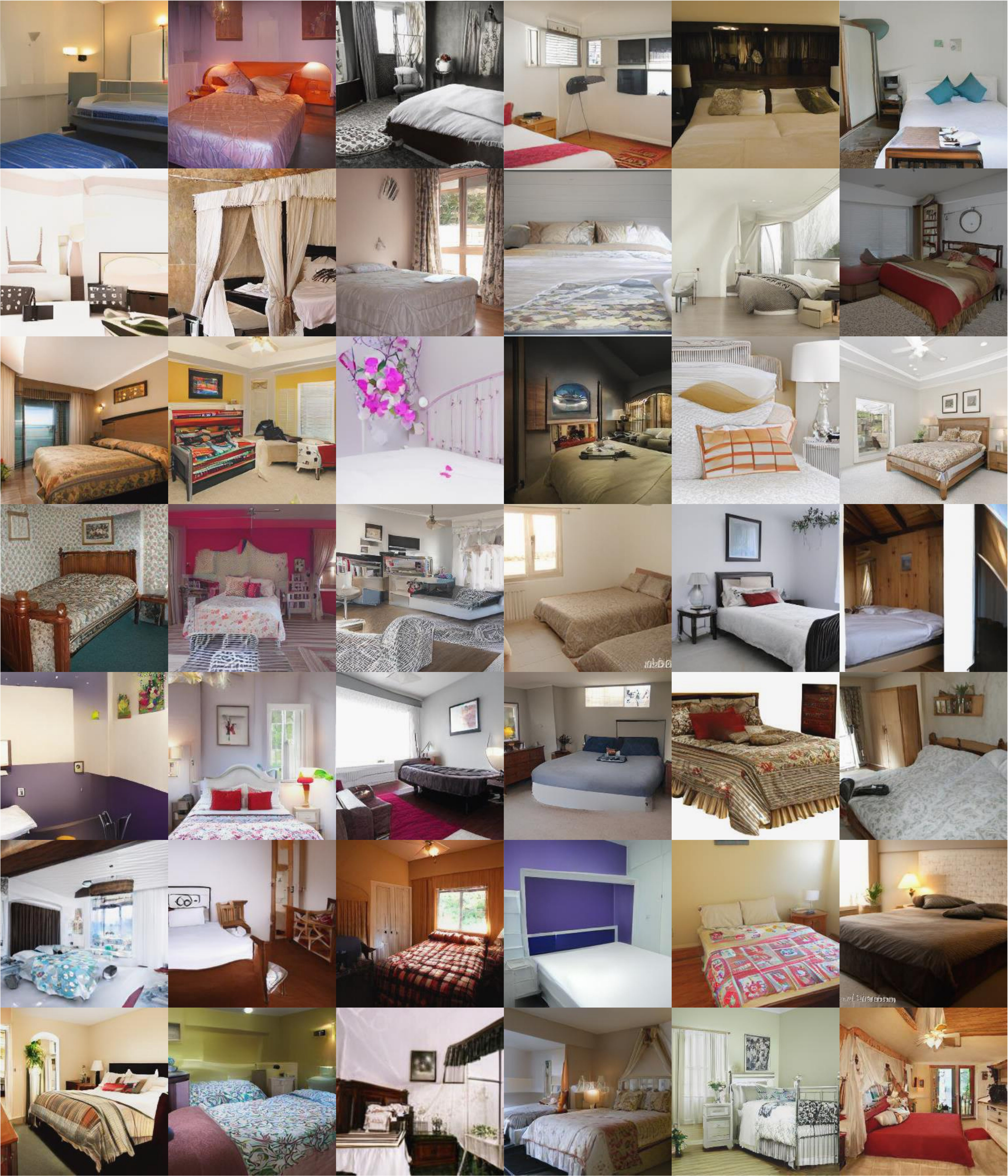}    \caption{\label{fig:uncurate_bedroom}Uncurated set of samples of SeQ-GAN+NAR on 256$\times$256 LSUN bedroom.}
\end{figure*}

We provide qualitative comparisons to BigGAN~\cite{brock2018large}, VQGAN~\cite{esser2021taming} and MaskGIT~\cite{chang2022maskgit} in \figref{fig:imagenet_cls0_1}, \figref{fig:imagenet_cls22_97} and \figref{fig:imagenet_cls108_141}. For MaskGIT and BigGAN, the samples are extracted from the paper and for VQGAN, we use their pre-generated samples in the official codebase\footnote{https://github.com/CompVis/taming-transformers}. Our SeQ-GAN+NAR produces results with better quality and diversity than previous methods. From the uncurated results in \figref{fig:uncurate_ffhq}, \figref{fig:uncurate_cat}, 
\figref{fig:uncurate_church} and
\figref{fig:uncurate_bedroom}, our SeQ-GAN+NAR can generate images with high quality and diversity on unconditional image generation.

\subsection{Limitation and Future Work}
\label{sec:limitation}

\subsubsection{Future Work}
Our observation 1 indicates that the quality of the discrete latent space in VQ-based generative models cannot be directly assessed by reconstruction fidelity, as the reconstruction and generation have different optimization goals. Thus, future work could design more intuitive methods to evaluate the quality of the discrete latent space.

In addition, observation 2 highlights the importance of semantics in the discrete latent space for visual synthesis. We have kept our approach simple by controlling the semantic ratio through the modification of the perceptual loss. Future works on VQ tokenizers can explore more effective ways to balance semantic compression and details preservation. For example, contrastive learning may improve the semantics compression of VQ tokenizers.

\subsubsection{Limitation}
\begin{figure*}[h]
    \centering
    \includegraphics[width=\linewidth]{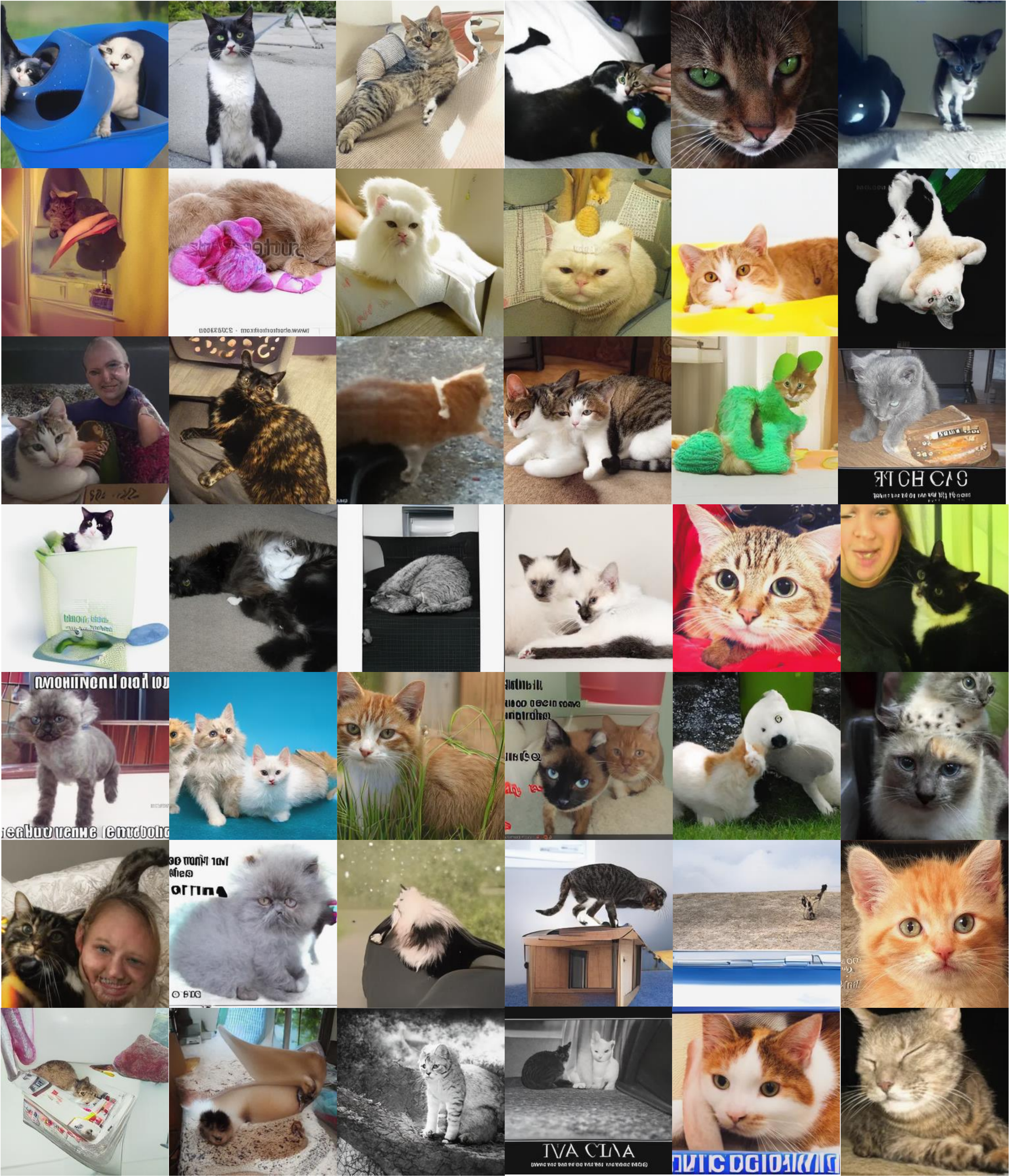}    \caption{\label{fig:uncurate_cat}Uncurated set of samples of SeQ-GAN+NAR on 256$\times$256 LSUN cat.}
\end{figure*}

Our method has the limitation inherited from likelihood-based generative models. While techniques such as rejection sampling~\cite{ramesh2021zero} and classifier-free guidance~\cite{ho2022classifier} can be used to filter out samples with bad shapes and improve sample quality in conditional image generation, there are few sampling techniques available for improving unconditional image generation. Classifier-based metrics such as FID tend to focus more on textures than overall shapes, and thus may not be consistent with human perception, as pointed out in \cite{geirhos2018imagenet}. To address this, StyleGAN2 introduces the perceptual path length (PPL) metric~\cite{karras2019style}, which is more related to the shape quality of samples. StyleGAN2 regularizes the GAN training to favor lower PPL. Although generative transformers with the SeQ-GAN tokenizer can achieve a better FID than StyleGAN2 in unconditional image generation, some samples still have poor overall shapes (as seen in the uncurated samples in \figref{fig:uncurate_cat}). Therefore, an interesting area for future research is to investigate the design of sampling techniques for unconditional image generation in likelihood-based generative models to improve overall shape quality.

\newpage

\newpage

\newpage

\newpage

\newpage

\newpage

\newpage

\end{document}